\documentclass[letterpaper, preprint, paper, 10pt]{AAS}	

\usepackage{bm}
\usepackage{amsmath}
\usepackage[colorlinks=true, pdfstartview=FitV, linkcolor=black, citecolor= black, urlcolor= black]{hyperref}
\usepackage{overcite}
\usepackage{footnpag}
\usepackage{caption}
\usepackage{subcaption}
\usepackage{longtable,tabularx}
\usepackage{tabu}
\usepackage{arydshln}
\usepackage{comment}
\usepackage{enumitem}
\usepackage{amsfonts}
\usepackage{multirow}
\usepackage[linesnumbered,ruled,vlined]{algorithm2e}

\PaperNumber{23-149}

\begin{document}

\title{Autonomous Guidance Navigation and Control\\
of the VISORS Formation-Flying Mission}

\author{
Tommaso Guffanti \thanks{Postdoctoral Scholar, Stanford University, Aeronautics and Astronautics, 496 Lomita Mall, Stanford, California 94305.},
Toby Bell \thanks{PhD Candidate, Stanford University, Aeronautics and Astronautics, 496 Lomita Mall, Stanford, California 94305.},
Samuel Y. W. Low \footnotemark[2],
Mason Murray-Cooper \footnotemark[2]
Simone D'Amico \thanks{Associate Professor, Stanford University, Aeronautics and Astronautics, 496 Lomita Mall, Stanford, California 94305.}}

\maketitle{} 		

\begin{abstract}
Virtual Super-resolution Optics with Reconfigurable Swarms (VISORS) is a distributed telescope mission for high-resolution imaging of the Sun using two 6U CubeSats flying in formation in a Sun-synchronous low-Earth orbit. An optics spacecraft carries a photon sieve acting as a high-resolution lens in the extreme ultraviolet spectrum, while the image passing through the sieve is focused on a detector spacecraft. This paper presents the newly conceived design of the on-board guidance, navigation and control (GNC) system, which is highly autonomous, robust, passively safe, and validated under realistic mission simulations. The primary objective of the GNC system is to establish a passively safe and high-precision formation alignment at 40-meter separation, with sub-centimeter relative navigation and position control accuracy, over repeated observations of 10-second duration. Science mission success rates are assessed via Monte-Carlo analyses under realistically modelled uncertainties stemming from sensing errors, maneuver errors, unmodelled dynamics, and erroneous knowledge of internal spacecraft components. Precise real-time relative navigation is achieved by carrier phase differential GPS with integer ambiguity resolution. Precise control over short baselines is achieved via closed-loop optimization-based stochastic model predictive control with centimeter-level accuracy. Control at far range and during approach is achieved by closed-form impulsive control with meter-level accuracy. Fault detection with isolation and recovery is constantly assessed by the GNC system, while passive safety is enforced throughout the mission to mitigate collision risks even under critical subsystem failure. Beyond VISORS, this work also realizes the crucial insight that the described GNC architecture is generalizable to other distributed space missions where accuracy and fault-tolerant safety are key requirements, such as rendezvous, proximity operations, and swarming missions.
\end{abstract}


\section{1. Introduction}
\label{section1}


\paragraph{Distributed space systems background}

Distributed space systems enable new mission concepts impossible for a single monolithic spacecraft, through the distribution of payloads, subsystems, and tasks over tailored inter-spacecraft geometries \cite{damico_pavone_miniat}. These concepts have enabled new science, such as precision gravimetry by the \textit{Gravity Recovery and Climate Experiment} (GRACE) mission \cite{GRACE, kirschner_2001_flight}, bi-static synthetic aperture radar imaging and digital elevation modelling by the \textit{TanDEM-X and TerraSAR-X} mission \cite{TDX, montenbruck_2008_navigation}, magnetometric mapping in highly eccentric orbits by NASA's \textit{Magnetospheric Multiscale} (MMS) mission \cite{MMS}, and, with strong relevance to distributed telescopy for this paper, precise coronagraphy by ESA's \textit{PROBA-3} mission \cite{Proba3_2013} and exoplanet detection and characterization using starlight suppression with starshades proposed by the \textit{Miniaturized Distributed Occulter Telescope} (mDOT) mission concept \cite{MDOT, MDOT2, mDOT_2019}. Distributed space systems also offer a promising testbed for advanced space technology demonstrations, such as the \textit{Canadian Advanced Nanospace eXperiment-4\& 5} (CanX-4\& 5) \cite{canx_07, canx_15}, the \textit{Prototype Research Instruments and Space Mission Technology Advancement} (PRISMA) mission \cite{PRISMA2012, gill_autonomous_2007}, the \textit{CubeSat Proximity Operations Demonstration} (CPOD) mission \cite{roscoe_2018_overview}, the \textit{Autonomous Vision Approach Navigation and Target Identification} (AVANTI) experiment \cite{gaias_autonomous_2014, gaias_avanti_2017}, the \textit{LUMELITE Formation-Flying Demonstration} mission \cite{goh2022optimized}, and the recently launched \textit{STARLING} mission \cite{STARLING}. Developments in distributed spacecraft missions have accelerated due to the availability of low size-weight-and-power (SWAP) commercial off-the-shelf (COTS) subsystems, sensors, and actuators. In this unique era, such developments enable high-impact cutting-edge science and engineering at low cost, as described in this paper.



\vspace{-5mm}
\begin{figure}[ht]
    \centering
    \includegraphics[width=0.65\linewidth]{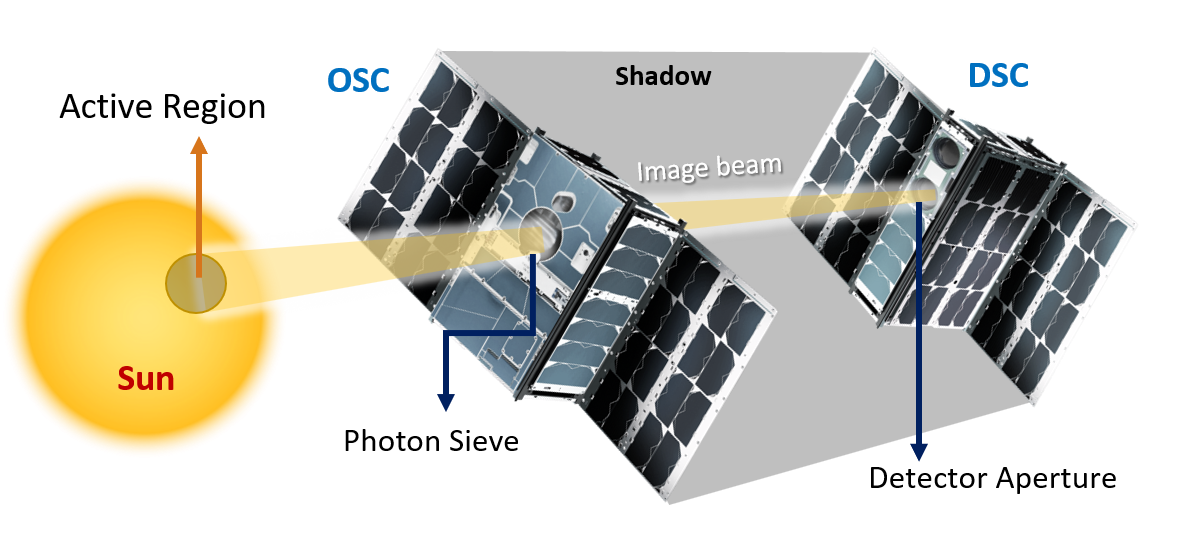}
    \caption{Notional diagram of VISORS in science mode (mock-up model, not to scale)}
    \label{fig:visors-illustration}
\end{figure}
\vspace{-5mm}

\paragraph{VISORS mission background}

Virtual Super-resolution Optics with Reconfigurable Swarms (VISORS) \cite{visors2021koenig, visors2022lightsey} is a distributed telescope mission consisting of two propulsive 6U CubeSats flying in formation in a Sun-synchronous low-Earth orbit. It is a multi-institution collaboration\footnote{\scriptsize University of Illinois Urbana-Champaign, Georgia Tech, NASA Goddard Space Flight Center, Stanford University, University of Colorado Boulder, Purdue University, New Mexico State University, Washington State University, Montana State University, Ohio State University and University of California San Diego.} first proposed during the National Science Foundation (NSF) IdeasLab 2018 \cite{NSFIdeaLab} to improve thermodynamic models of the solar corona through high-resolution imagery. The science objective of VISORS is to image active regions of the Sun's corona in the extreme ultraviolet with high resolution. To achieve this, the optics spacecraft (OSC) carries a photon sieve payload that focuses light via constructive interference on a detector payload mounted on the detector spacecraft (DSC), creating a high-contrast image with 0.1-arcsecond resolution. Observations occur once per orbit at 40-meter separation with 10-second duration. Precise line-of-sight alignment between the active region, OSC, and DSC requires sub-centimeter relative navigation and control accuracy with passive collision safety. The full mission requirements and subsystems are detailed in \hyperref[section2]{Section 2}. The Stanford Space Rendezvous Laboratory (SLAB) is chiefly responsible for the design and development of the autonomous VISORS GNC system described in this paper. VISORS is due for launch in October 2024 on SpaceX Transporter-12.


\paragraph{Relevant distributed GNC architectures}

A broad survey of GNC requirements for formation-flying spacecraft missions is given by Di Mauro et al \cite{di2018survey}. The uniqueness of the VISORS mission GNC architecture is characterized by its highly precise (cm-level) navigation and control, enacted by a fully autonomous GNC system and enabled by low-cost COTS components. Similar but less demanding GNC architectures for SmallSats and NanoSats are found in \textit{PRISMA} \cite{PRISMA2012} (2010), \textit{CAN-X 4/5} \cite{canx_15, canx_07} (2014), and \textit{CPOD} \cite{roscoe_2018_overview} (2022), which also leverage COTS components with low SWAP while meeting high navigation and control accuracy requirements.

The PRISMA GNC architecture for the SAFE experiment \cite{prisma2012safe, damico_phd_2010} integrated five key blocks: data interfacing, orbit determination, orbit prediction, formation reconfiguration, and formation station-keeping. The relative navigation filter employed on-board real-time carrier phase differential GPS (CDGPS) with estimation of carrier phase float ambiguities in an extended Kalman filter. Relative orbit control applied closed-form impulsive maneuvers for reconfiguration. PRISMA demonstrated relative navigation accuracy of 5 cm (position) and 1 mm$\cdot$s\textsuperscript{-1} (velocity) in 3D RMS, and 1-meter relative orbit control accuracy at a closest approach of 100 m.

CanX-4/5 achieved similar performance despite its smaller form factor of 20 cm\textsuperscript{3}. CanX-4/5 demonstrated successful on-board real-time relative navigation with 10 cm accuracy also using CDGPS with float ambiguity estimates; it achieved sub-meter relative orbit control using a discrete-time linear quadratic regulator (LQR) for formation station-keeping and impulsive maneuvers optimized over energy-like cost functions for formation reconfiguration. CanX-4/5 achieved a closest approach of 50 m.

The CPOD GNC system, while ultimately unsuccessful in full deployment, proposed a comprehensive suite of control algorithms and metrologies in its navigation filter. CDGPS measurements are used with ranges from an intersatellite link and bearing angles from a vision-based sensor \cite{roscoe_2018_overview} in the filter update. An image processing block assigned bearing angle measurements to targets. Formation control was split into three blocks: formation reconfiguration employed sequences of $n$-impulse fuel-optimal maneuvers \cite{roscoe_2015_formation}; formation maintenance employed model predictive control \cite{gvecontrol2007}; rendezvous and approach adopted a continuous thrust local-vertical local-horizontal (LVLH) state feedback control law. Each control block was tailored to the desired precision required in each mission phase. CPOD's GNC system also sported an autonomous safety block which monitored conjunction assessment, fault detection, correction, and an abort maneuver logic. While CPOD deployed an extremely comprehensive GNC architecture, it was unable to fulfil the rendezvous, proximity operations and docking (RPOD) objectives due to system level challenges \cite{NASA_CPOD_2023}.


\paragraph{Contributions of this work}

The VISORS GNC system advances the state of the art on five fronts. First, the navigation filter employs CDGPS measurements with integer ambiguity resolution (IAR) to meet mm-level relative navigation accuracy requirements, using SLAB's Distributed Multi-GNSS Timing and Localization (DiGiTaL) software. To the best of the authors' knowledge, in-orbit real-time IAR has never been achieved on CubeSats. VISORS presents an opportunity to demonstrate this first-ever milestone in distributed satellite navigation. Furthermore, inclusion of a laser rangefinder (LRF) presents another opportunity to demonstrate on-orbit sensor fusion between range and CDGPS with IAR \cite{low2024iarcoupled}. Second, the VISORS GNC system boasts a unique suite of impulsive control algorithms, including convex optimization-based control solutions formalized using reachable-set theory \cite{koenig_2021_opt} that will achieve first-ever flight heritage on VISORS and closed-form impulsive control solutions \cite{gaias_imp_2015, chernick_2018_closed} with partial flight heritage \cite{gaias_avanti_2017}. Optimization-based control is achieved in embedded flight software using the Embedded Conic Solver (ECOS) \cite{ecos_13}, an interior-point solver for second-order cone programming (SOCP) \cite{boyd_book}. In-space convex optimization-based translational control is a novelty for CubeSats and a contribution by VISORS. To the authors knowledge, the only other attempt at convex optimization on CubeSats has been for optimal attitude control \cite{SOCI_cubesat}. Third, VISORS will demonstrate a fully autonomous fault-tolerant safety concept allowing operations at extremely close range while using low-cost COTS components. This concept combines passively-safe orbit design \cite{damico_proximity_2006, guffanti_jgcd_2023}, autonomous collision-detection, and reactive optimal escape maneuvers. Fourth, the VISORS GNC software demonstrates use of a modern software development approach to integrate navigation, control, and safety blocks into a single fully autonomous GNC system that minimizes memory and computational cost while delivering unprecedented navigation and control accuracy. Fifth, the VISORS GNC flight software performance is validated through thorough high-fidelity Monte Carlo analyses of multiple mission scenarios, incorporating realistic uncertainties in measurements, control actuation, unmodeled dynamics, and errors in the knowledge of internal spacecraft components. This analysis demonstrates the state-of-the-art capabilities of the GNC system and testifies to its importance as a mission-enabling technology for VISORS.

Beyond VISORS, this work also realizes the crucial insight that the described GNC architecture may be generalizable to other distributed space missions where accuracy and fault-tolerant safety are key requirements, such as rendezvous, proximity operations, and swarming missions. Such a generalized high-precision GNC architecture would likely be unmatched in capabilities by any existing CubeSat-class mission.


\section{2. Mission requirements}
\label{section2}


\subsection{Science requirements}

\begin{figure}[h]
    \centering
    \includegraphics[width=0.95\linewidth]{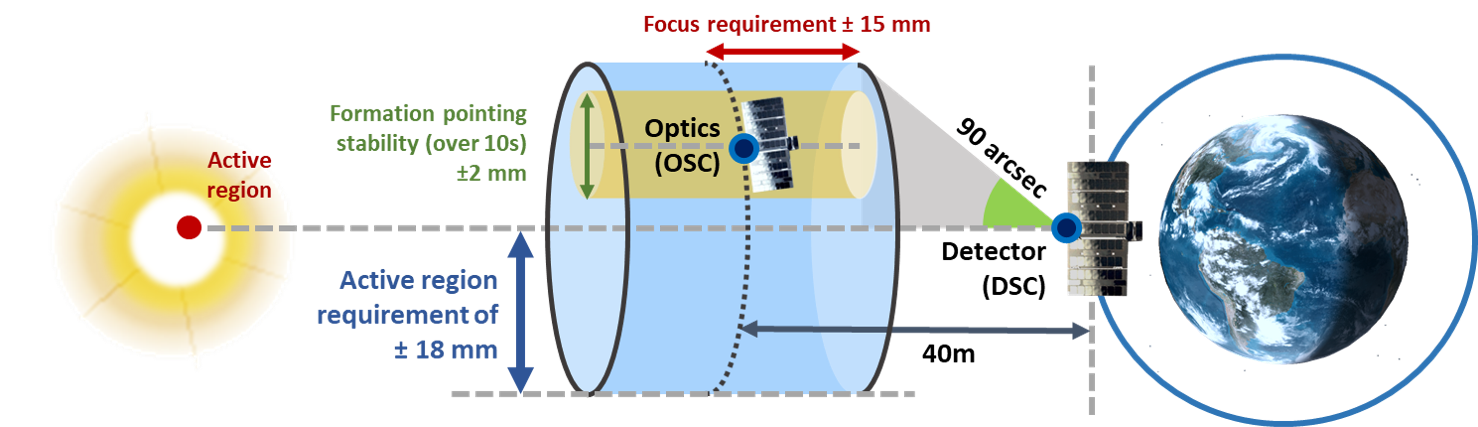}
    \caption{VISORS alignment requirements in science mode (not to scale)}
    \label{fig:visors-requirements}
\end{figure}

During a science observation, the OSC photon sieve is in line-of-sight alignment with the DSC detector and the target active region (see Figure \ref{fig:visors-requirements}) to create a high-contrast image with a 0.1" resolution in the extreme ultraviolet. The observation window is 10 s long. The science observation requirements on image quality (pointing, stability, drift, and focus) translate into the following three geometric alignment requirements:

\begin{itemize}[noitemsep,topsep=0pt]
    \item {\it Alignment with active region (pointing)}: The center of the photon sieve must not deviate from the line connecting the center of the detector aperture and the target active region by more than 18 mm.
    \item {\it Line-of-sight stability/image drift}: The spacecrafts' inertial relative velocity in the plane perpendicular to the line of sight must not exceed 0.2 mm/s, so common features can be tracked across exposures.
    \item {\it Image focus}: The distance between the detector aperture center and the photon sieve pattern center must not deviate by more than 15 mm from the nominal separation of 40 m.
\end{itemize}

These three critical requirements were intended so that meeting 1 successful observation out of 5 attempts ($20\%$) is probabilistically sufficient to declare mission success.

The key risks and challenges associated with the execution of such high-precision close-proximity alignment stem from VISORS leveraging multiple low-cost COTS with no flight heritage, including the propulsion and inter-satellite crosslink systems. This mandates a GNC system with unparalleled precision performance in navigation in control, but also necessitates fault-tolerant functionalities for all phases of the mission. The following sections details further the safety and bus-derived requirements.


\subsection{Safety requirements}

Safe separation of at least $\epsilon$ shall be enforced even in case of sudden loss of control capabilities, i.e. passive safety \cite{damico_proximity_2006, guffanti_jgcd_2023}, and checked on-board every $n_{S}$ seconds over a $T$-orbit horizon at $q$-$\sigma$ confidence. The parameters $\epsilon$, $n_s$, $T$, and $q$ can be tuned during the mission to regulate the responsiveness of the safety system. Nominally, safe separation is checked and enforced in the radial-normal (RN) plane, to be robust to uncertainties in the relative along-track position, which grow unboundedly through Keplerian dynamics. A less conservative approach is checking 3D separation, which can be used if neccessary. Furthermore, whenever the $q$-$\sigma$ passive safety margin is violated, the formation must be separated in either along-track or RN-plane through escape maneuvers. This combined passive and reactive approach mitigates collision risk even in case of critical subsystem failure and bus safe modes. In Section 4, Table \ref{tab:gnc-algos} lists the safety parameters used for VISORS. In addition, the GNC system shall be robust to internal failures and mitigate them either internally or by interaction with the spacecraft fault detection, isolation, and recovery (FDIR) logic. Section 4 describes the GNC FDIR and how it interfaces with the higher-level spacecraft FDIR. Despite its distribution across two satellites, the mission still remains single-failure intolerant. For example, a prolonged uncontrolled safe mode simultaneously on both spacecraft remains an irreducible risk. These risks have been assessed during the critical design phases of the mission and deemed acceptable for this mission class.


\subsection{Spacecraft systems requirements and design}

\paragraph{Sensors}
Each spacecraft employs a Blue Canyon Technologies (BCT) Nano Star Tracker (NST) and Sun Sensor, with the DSC employing an additional NSTs (2 in total on DSC) to maximize attitude stability. L1-only pseudorange and carrier phase GPS measurements are received through a Novatel OEM719 receiver and antenna. Sensor measurements are exchanged across a near-omni-directional inter-satellite crosslink with $\leq$ 10 km range. The GPS antennas for both spacecraft are required to point within 30$^\circ$ of the zenith direction for at least one orbit before and after any observation to maximize the signal-to-noise ratio and the number of commonly visible GPS satellites. Finally, a laser rangefinder on the OSC provides an unambiguous confirmation of payload alignment between the DSC and OSC, while also providing additional metrology for the relative navigation filter.

\begin{figure}[ht!]
    \centering
    {\includegraphics[width=0.45\textwidth]{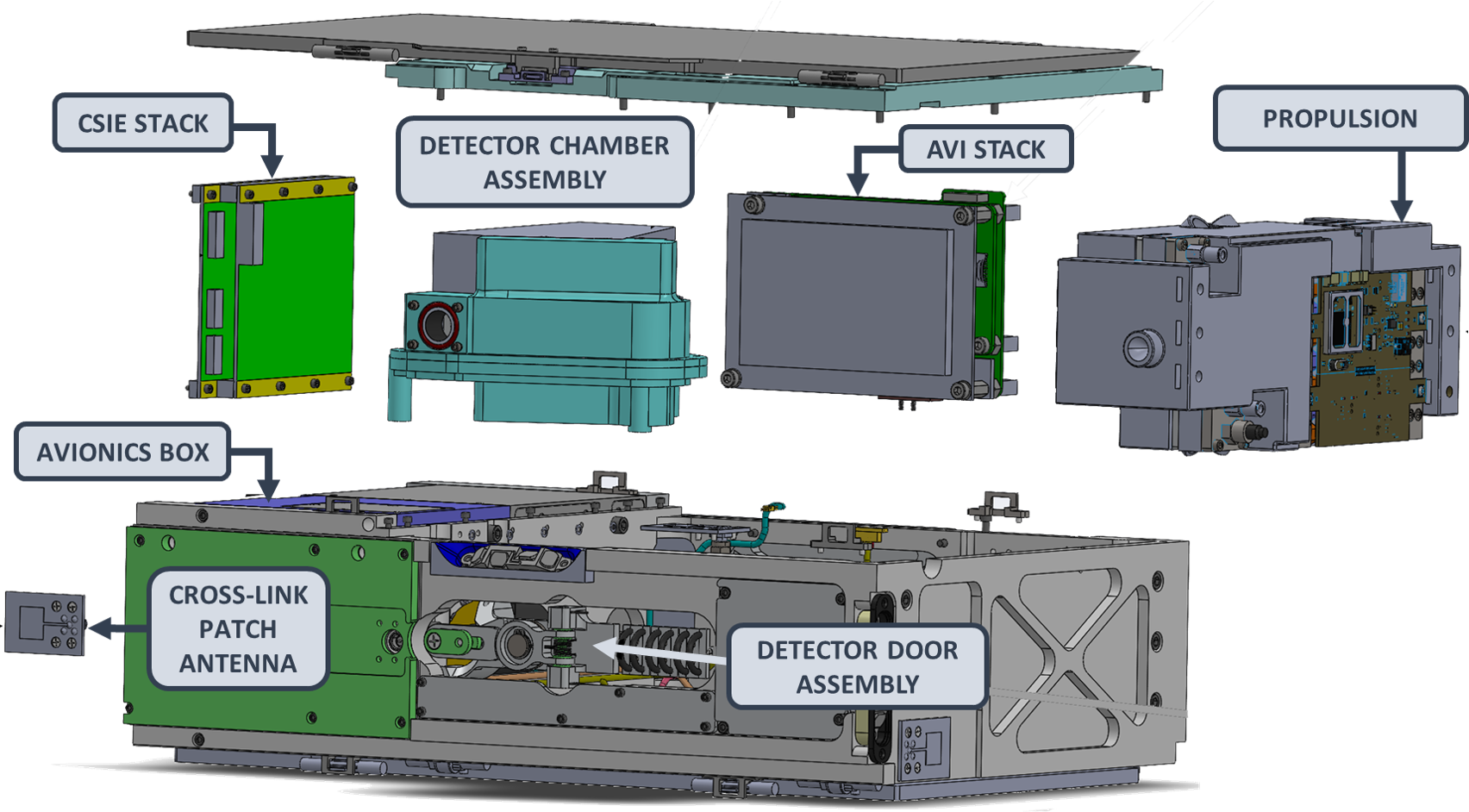}} \hspace{0.3cm}
    {\includegraphics[width=0.45\textwidth]{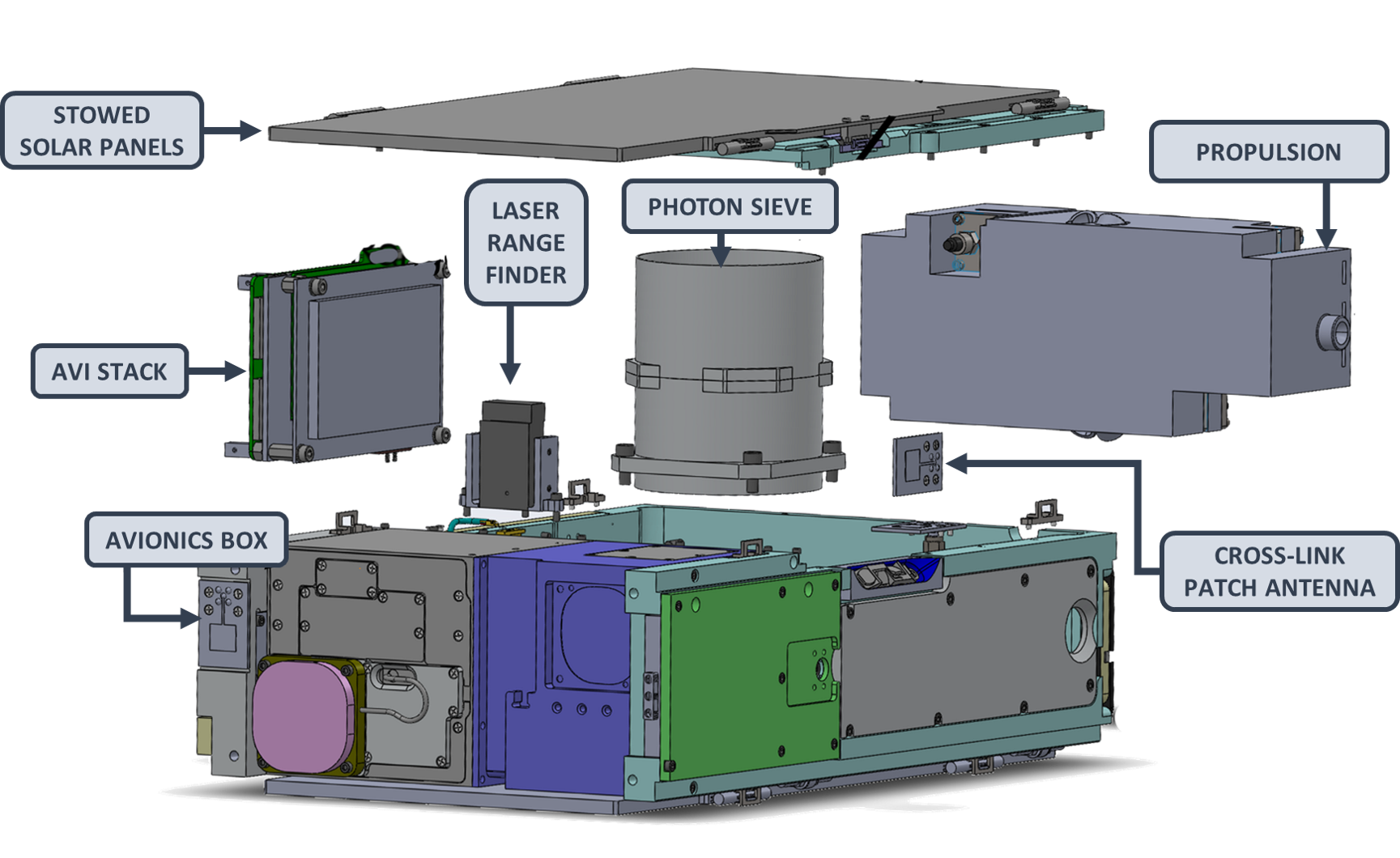}}
    \vspace{3mm}
    \caption{DSC (left), OSC (right); bus by Blue Canyon Technologies, payload by the project team \cite{visors2022lightsey}}
    \label{fig:visors-model}
\end{figure}

\paragraph{Actuators}
The proposed GNC architecture is primarily concerned with translational (relative orbit) control. Attitude actuation is managed outside of this GNC architecture at the bus-level, using $03 \times$ BCT's RWP050 reaction wheels (0.050 Nms and 0.007 Nm of momentum and max torque), and 0.6 Am\textsuperscript{2} torque rods. A 3D-printed cold-gas propulsion system \cite{lightsey_18} provides $\sim$14.6 m/s of $\Delta \boldsymbol{v}$ on the DSC and $\sim$8.4 m/s on the OSC. It has 6 nozzles (3 opposing pairs on perpendicular axes), ensuring the spacecraft can execute arbitrary maneuvers without changing attitude. Propulsion can actuate maximum impulses of 0.023 Ns ($\sim$2 mm/s), and requires 46.5 s after each maneuver to refill the plenum (Table \ref{tab:visors_act}). As described in Section 4, this requires a maneuver-splitting capability within GNC to decompose maneuvers larger than 2 mm/s and spread them around the orbit so they can be realized by the available actuation. In addition, the actuation system introduces uncertainty in maneuvers' magnitude and direction (Table \ref{tab:visors_act}). This uncertainty affects control performance and is assessed through Monte Carlo analysis in Section 5.

\begin{table}[ht!] 
\centering
\caption{Assumed spacecraft system parameters}
\resizebox{\linewidth}{!}{
\subfloat[Spacecraft geometry uncertainty]{
  \label{tab:visors_int_comp}
  \begin{tabular}{l c}
    \hline 
    \hline
    Feature & 1-$\sigma$ \\
    \cline{1-2}
    Center of mass variation & 3 mm \\
    GPS antenna phase center offset\textsuperscript{$\dagger$} & 8.54 cm \\
    Mounting position error\textsuperscript{*} & 0.5 mm \\
    Mounting direction error\textsuperscript{*} & 30 arcsec \\
    \hline
    \hline
    \\
    \textsuperscript{*} For the optical payloads, LRF
    \\
    \textsuperscript{$\dagger$} From the static base of the antenna mount
  \end{tabular}
}
\subfloat[Propulsion system performance\cite{lightsey_18}]{
  \label{tab:visors_act}
  \begin{tabular}{ l c c }
    \hline 
    \hline
    Parameter & Value & 1-$\sigma$ \\
    \cline{1-3}
    Minimum impulse (Ns) & 1$\times$10\textsuperscript{-3} & – \\
    Maximum impulse (Ns) & 2.2$\times$10\textsuperscript{-2} & – \\
    Impulse quantization (Ns) & 6$\times$10\textsuperscript{-5} & – \\
    Maneuver angle error (deg) & 0 & 0.05--10 \\
    Maneuver magnitude error (\%) & 0 & 5--50 \\
    Specific impulse (s) & 44 & 2 \\
    Max time to refill plenum (s) & 45 & – \\
    Max time to empty plenum (s)  & 1.5 & – \\
    Maneuver spread (s) & 46.5 & – \\
    \hline
    \hline
  \end{tabular}
}
}
\end{table}

\paragraph{Spacecraft internal components}

\begin{figure}[ht!]
    \centering
    \includegraphics[width=0.75\linewidth]{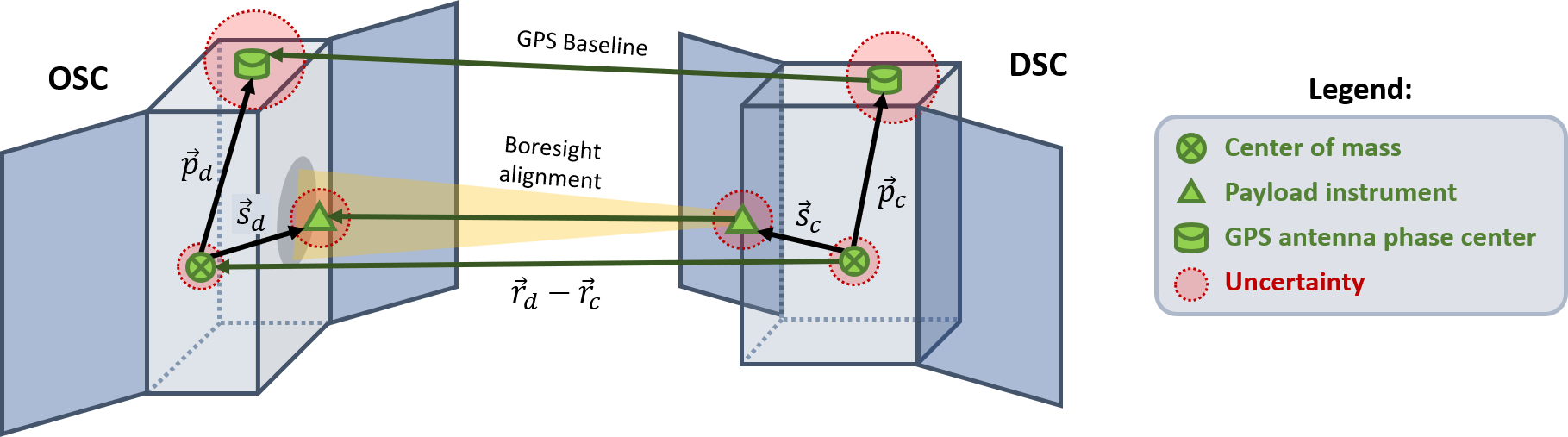}
    \caption{Spacecraft internal components uncertainty}
    \label{fig:visors-int-comp}
\end{figure}

Different reference points are used in navigation, control, and observation alignment. The GPS measurements in navigation refer to the GPS antennas' electrical phase center. The dynamics models used in navigation and control relate to the true centers of mass. The observation success relates to the alignment of the optical payloads (Figure \ref{fig:visors-int-comp}). It is assumed that the optical payload's body-frame coordinate is constant. Assumed errors in the true center of mass and GPS phase center are provided in Table \ref{tab:visors_int_comp}. The static center of mass is a known body-frame coordinate from the 3D model. However, the true center of mass actually varies and deviates away from the static center of mass due to fuel depletion, fluid slosh, and solar panel orientation. The position of the GPS antenna phase center offset is modelled as an along-boresight bias, influenced by the choice of antenna, ground plane, frequency, and multi-path effects. It is expected that the phase center offset would have a larger uncertainty than the variation of the true center of mass. The relative position between the electrical phase center of the GPS antenna to the static center of mass of both spacecraft, $\Vec{p}_c$ and $\Vec{p}_d$, are estimated by the navigation filter in real time.

\section{3. Concept of operations and orbit design}
\label{section3}

\begin{figure}[ht!]
    \centering
    \includegraphics[width=\linewidth]{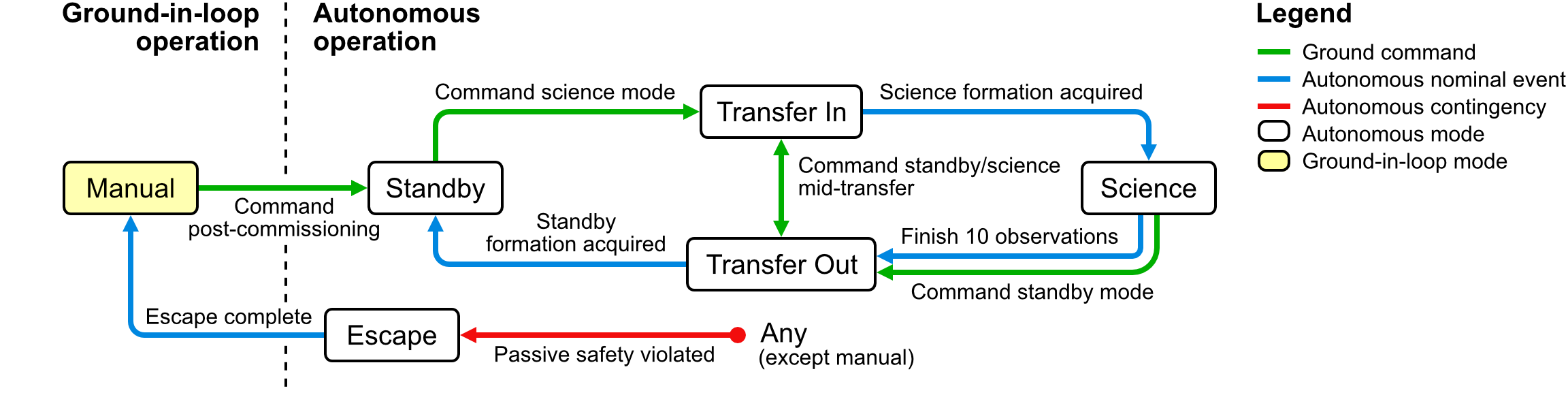}
    \caption{VISORS primary mission modes}
    \label{fig:visors-modes}
\end{figure}

\begin{figure}[ht!]
    \centering
    \includegraphics[width=0.55\linewidth]{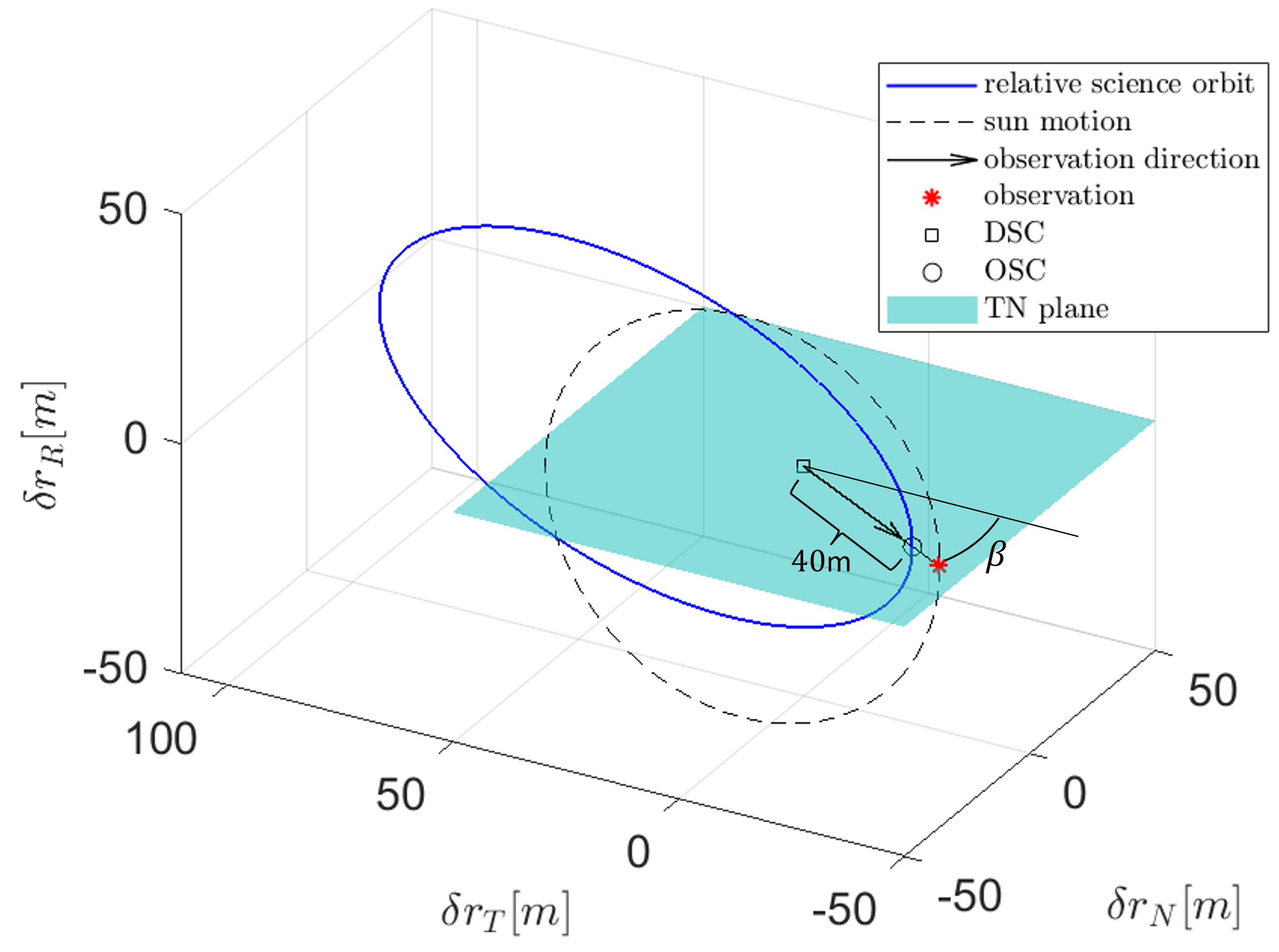} \hspace{0.2cm}    \includegraphics[width=0.31\linewidth]{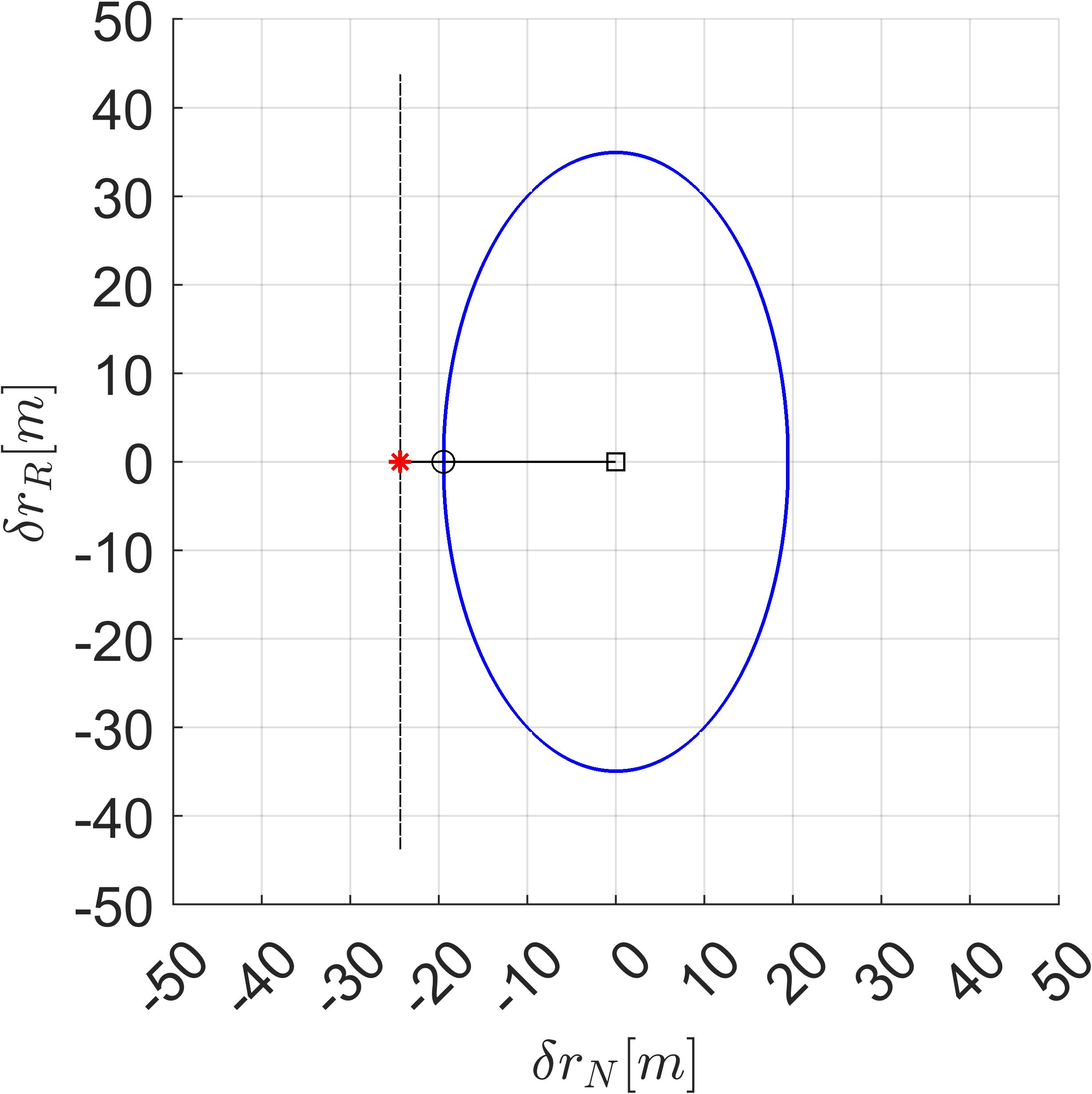} \\ \vspace{0.3cm}\includegraphics[width=0.40\linewidth]{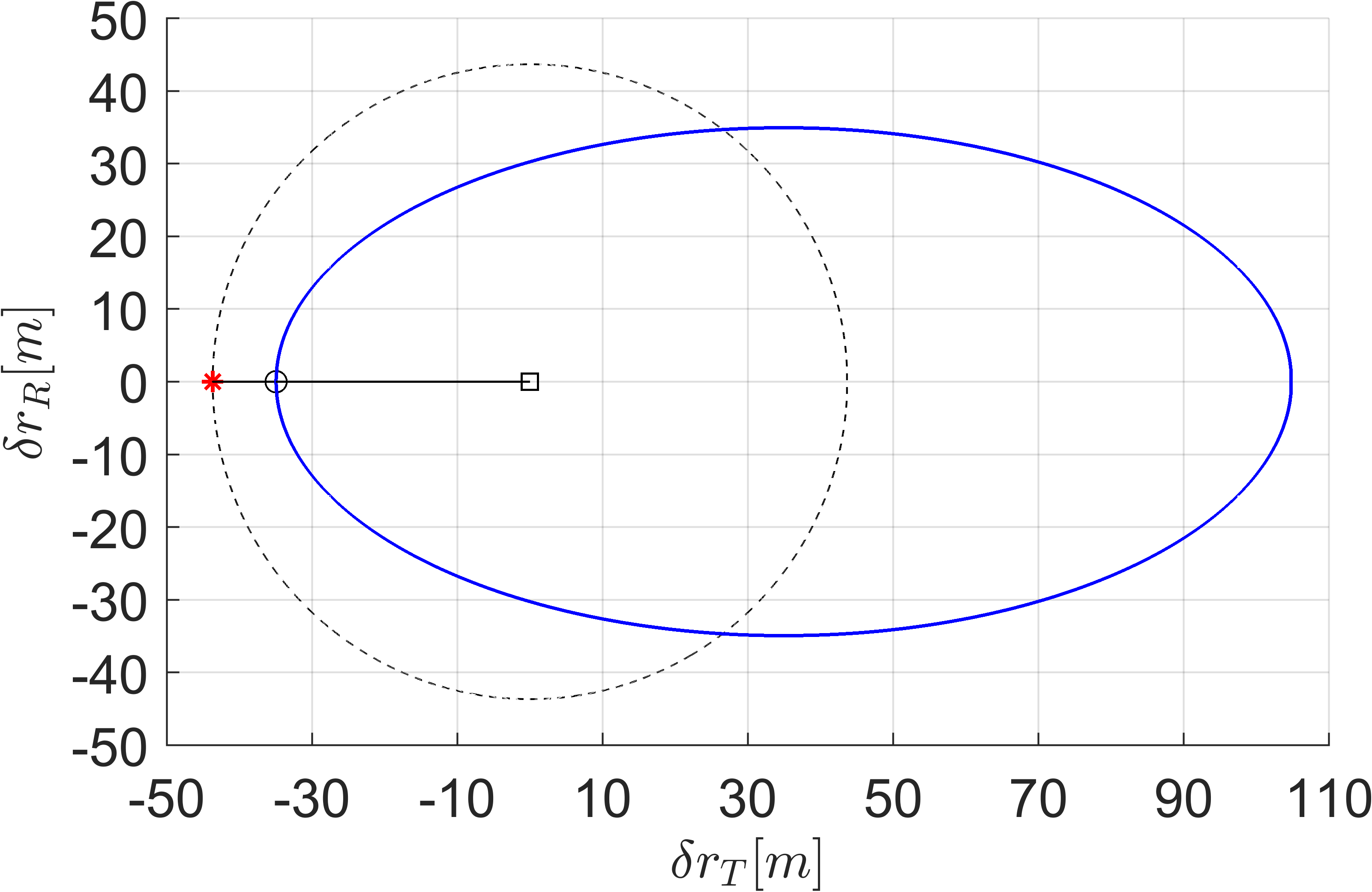} \hspace{0.2cm} \includegraphics[width=0.5\linewidth]{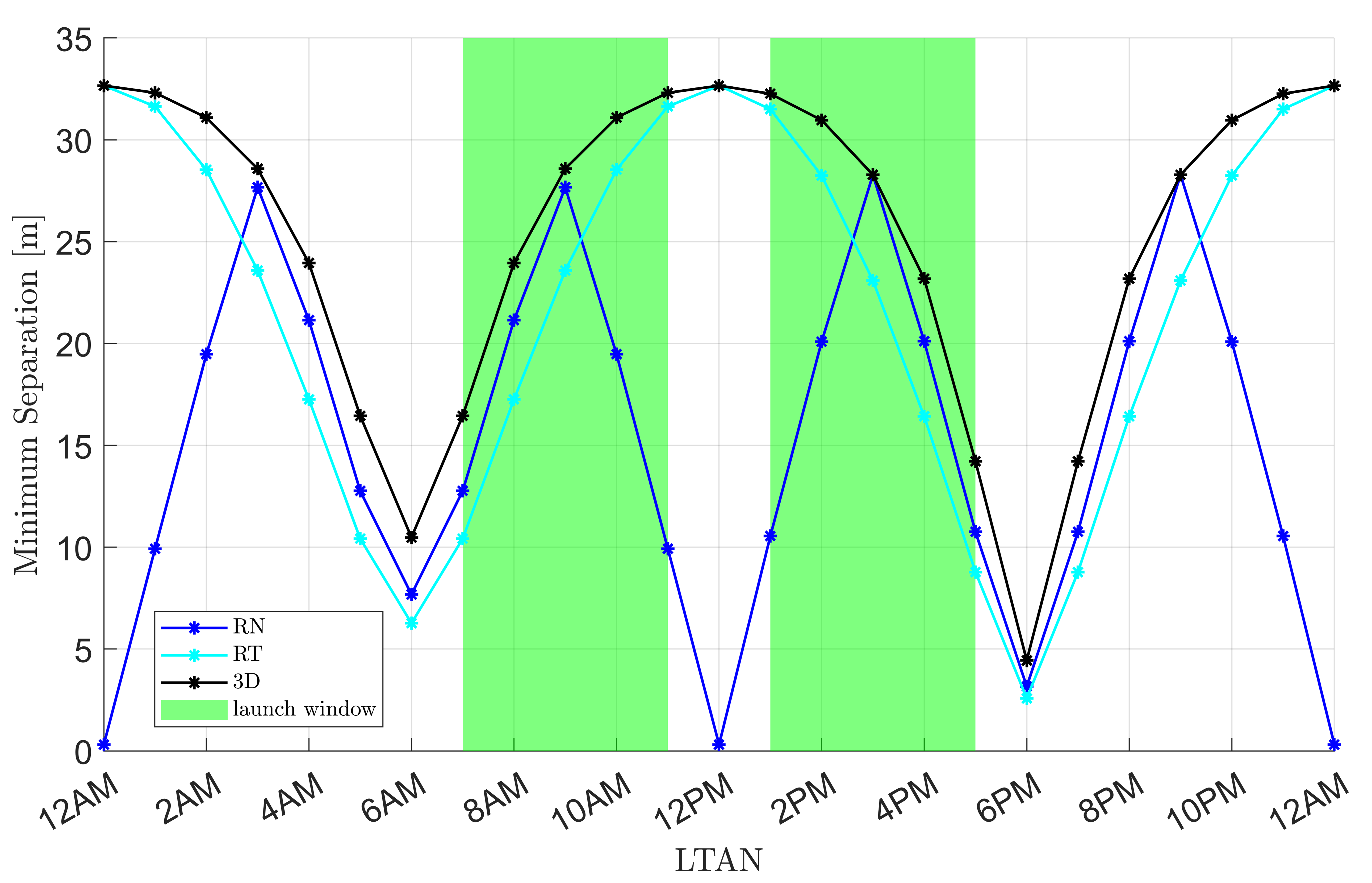} \\
    \caption{Science mode relative orbit. For LTAN = 10am, 3D RTN trajectory (top-left), RN and RT projections (top-right, and bottom-left). Varying the LTAN, minimum separation in 3D, RN and RT (bottom-right, perturbations and uncertainty effects neglected).
    }
    \label{fig:visors-observations}
\end{figure}

The VISORS concept of operations and orbit design are reviewed briefly here; a detailed elaboration is given by Lightsey et al \cite{visors2022lightsey} and Koenig et al \cite{visors2021koenig}. Mission operations are divided into five key modes (Figure \ref{fig:visors-modes}):

\begin{itemize}[]
    \item {\it Manual mode}: No autonomous control is performed, only ground-commanded maneuvers. Used at the beginning of the mission for initial checkout and formation acquisition, in contingency scenarios until nominal operations are restored, and at the end of the mission for decommissioning.
    \item {\it Standby mode}: Used for downlinking science data and waiting for science opportunities. Entered by command from manual mode or autonomously from transfer mode. The formation may remain in standby mode for up to a few weeks. The standby relative orbit is passively safe with a minimum RN-plane separation of 200 m, achieved via relative eccentricity–inclination separation \cite{damico_proximity_2006}.
    \item {\it Science mode}: collects images of the Sun's target active region once per orbit. Entered autonomously from transfer mode, and exited autonomously to transfer mode after the commanded number of observation attempts (nominally 10). To fullfill the relative acceleration and velocity requirements, the observations are performed once per orbit close to the TN-plane \cite{visors2021koenig}. The science mode relative orbit is nominally passively safe with a minimum RN-plane separation achieved via relative eccentricity–inclination separation \cite{damico_proximity_2006}. Figure \ref{fig:visors-observations} presents the geometry of a possible science relative orbit, with the DSC at the origin of the radial-tangential-normal (RTN) frame. The axes represent the relative spacecraft position components in the RTN frame (in 3D on the top-left, and the RN and RT projection on the top-right and bottom-left). The science relative orbit configuration required to have alignment with the sun on the TN-plane depend on the orbit $\beta$-angle (i.e., the angle between the pointing vector to the target region on the Sun and its projection onto the orbit plane) \cite{visors2021koenig}. The minimum separation on such relative orbit can be expressed as a function of the $\beta$-angle, implying a constraint on the local time of the ascending node (LTAN) and altitude of the launched Sun-synchronous orbit to have passively safe science operations. Figure \ref{fig:visors-observations} (bottom-right) shows the minimum 3D, RN and RT separations on the nominal science relative orbit as a function of the launched LTAN, the green area correspond to the planned launch window for the Transporter-12 rocket. 
    \item {\it Transfer mode}: reconfigures the formation between the standby and science relative orbits. This mode is entered from standby mode by ground command, or autonomously from science after completing the commanded number of observations attempts. This mode exits autonomously upon completing the reconfiguration to the science or standby orbit. Passive safety is maintained not only in the standby and science modes' relative orbit designs, but also at every instant of the transfer. This requires planning for passive safety after each intermediate maneuver execution.
    \item{\it Escape mode}: entered autonomously during any mission phase when GNC detects a passive safety violation, either in RN or in $3$D if the LTAN does not permit a safe minimum RN separation. It maneuvers to increase spacecraft separation and exits autonomously to manual mode.
\end{itemize}

\section{4. GNC architecture and algorithms}
\label{section4}

\subsection{4.1. Architecture}

\begin{figure}[ht!]
    \centering
    \includegraphics[width=0.95\linewidth]{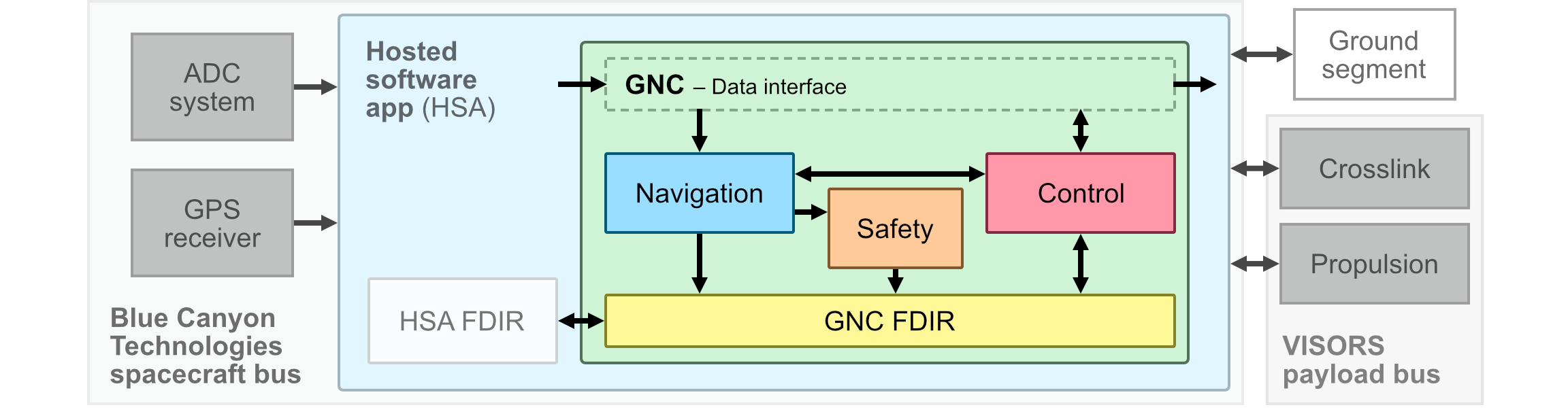}
    \caption{Component-level VISORS GNC architecture}
    \label{fig:gnc-architecture}
\end{figure}

This paper presents an overview of VISORS's GNC architecture (Figure \ref{fig:gnc-architecture}) and the novel algorithms powering each GNC subsystem (Table \ref{tab:gnc-algos}). The GNC system runs as a subcomponent of the hosted software app (HSA) on the bus flight computer. It communicates with hardware systems such as attitude determination and control (ADC), crosslink, propulsion, sensors such as the GPS receiver and LRF, software subsystems such as the HSA FDIR, and the ground segment through commands and telemetry. GNC internally includes navigation, control, safety, and GNC FDIR modules supported by data interfacing logic.

\paragraph{Interfaces} The GNC component interfaces with the hosted software app (HSA) through a specification of input and output channels (Table \ref{tab:io}). Each channel has a memory layout and purpose known to GNC and the hosted software app. All channels have a nominal rate, but GNC is asynchronous by design and can receive inputs on any channel at any time. Similarly, GNC is robust to missing inputs and can function in the absence of inputs on any channel, albeit with reduced performance when missing critical inputs like GPS messages.

All inputs and outputs flow through the HSA; GNC does not interface directly with hardware. However, most of GNC's input/output channels imply communication with a particular hardware subsystem. For example, in Table \ref{tab:io}, the 1-second GPS input corresponds to the HSA receiving data from the on-board GPS receiver and forwarding it to GNC. Similarly, GNC expects that the propulsion system sends telemetry (e.g., health, tank pressure) after every maneuver, at most every 46.5 s. However, the details of hardware communication are left to the HSA, implemented by the Space Systems Design Lab at Georgia Tech.

\begin{table}[ht!] 
\centering
\caption{VISORS GNC inputs and outputs}
\label{tab:io}
\resizebox{0.8\linewidth}{!}{
\begin{tabular}{l c c l c}
  \hline
  \hline
  Input & Nominal period & & Output & Nominal period \\
  \cline{1-2} \cline{4-5}
  Configuration parameters & 1 d && Maneuver & 46.5 s \\
  Time at Tone & 1 s && Star tracker power & 45 min\textsuperscript{\dag} \\
  GPS message & 1 s && LRF power & 45 min\textsuperscript{\dag} \\
  LRF measurement & 2 s\textsuperscript{*} && Telemetry & 1 s \\
  Crosslink message & 1 s && Crosslink message & 1 s \\
  Ground command & 1 week && Observation slew & 90 min\textsuperscript{\dag} \\
  Propulsion telemetry & 46.5 s && Observation start & 90 min\textsuperscript{\dag} \\
  GNC target & 1 week && Observation end & 90 min\textsuperscript{\dag} \\
  \cline{1-2}
  &&& Current mode & 1 d \\
  \textsuperscript{*} only during observation &&& GNC incapable & – \\
  \textsuperscript{\dag} only in science mode &&& Passive safety violated & – \\
  \cline{4-5}
\end{tabular}}
\end{table}

\paragraph{Algorithms} The GNC library includes a unique suite of navigation, control, and collision detection and avoidance algorithms (Table \ref{tab:gnc-algos}). Navigation leverages carrier phase differential GPS with on-board real-time integer ambiguity resolution (IAR), in the form of Stanford’s Distributed Multi-GNSS Timing and Localization (DiGiTaL) software package. When IAR is on, it provides sub-cm relative position and sub-mm/s relative velocity estimates. For autonomous formation control, impulsive maneuver planning is used in conjunction with collision detection for safety. In particular, before science observations, a novel convex optimization-based stochastic model predictive control strategy achieves the required sub-cm alignment accuracy. During far-range and approach phases, closed-form impulsive control solutions with partial flight heritage allow for meter-level accurate control. For fault tolerance, safe separation even in case of sudden loss of control capabilities, a.k.a. passive safety, is enforced in the RN-plane by each control algorithm throughout the mission. Even so, safe separation is checked numerically every 10 s, and when it is found violated, a novel analytical single-impulse escape solution provides immediate RN-plane separation and along-track drift.

\begin{table*}[ht!] 
\centering
\caption{VISORS GNC algorithms \label{tab:gnc-algos}}
\resizebox{\linewidth}{!}{
\begin{tabular}{c r l c l r l c c c r l}
  \hline
  \hline
  \multicolumn3c{DGPS navigation} && \multicolumn4c{Impulsive control} && \multicolumn3c{Safety} \\
  \cline{1-3} \cline{5-8} \cline{10-12}
  Algorithm & \multicolumn2c{Accuracy (3D 1-$\sigma$)} & & Algorithm & \multicolumn2c{Features} & \begin{tabular}{c} Horizon \\ Accuracy \end{tabular} & & Type & \multicolumn2c{Features} \\
  \cline{1-3} \cline{5-8} \cline{10-12}
  DiGiTaL & Abs. & 5 m, 1 cm/s && Transfer \& tracking & Man. type & 3 T, 1–2 N & \textgreater 1.5 orb. && \multirow4*{Passive} & Horizon $T$ & 1–2 orb. \\
  (IAR on)\cite{giralo_phd_2021} & Rel. & 1 cm, 50 \textmu m/s && (analytical)\cite{gaias_imp_2015, chernick_2018_closed} & Man. freq. & 0.5 orb. & 1–10 m &&& Plane & RN or 3D \\
  \cdashline{1-3} \cdashline{5-8}
  DiGiTaL & Abs. & 5 m, 1 cm/s && Escape & Man. type & 1 RTN & ASAP &&& Min. sep. $\epsilon$ & 5 m \\
  (IAR off)\cite{gill_autonomous_2007} & Rel. & 5 cm, 100 \textmu m/s && (analytical) & Man. freq. & – & 1–10 m &&& Confidence $q$ & 2-3 $\sigma$ \\
  \cline{1-3} \cdashline{5-8} \cdashline{10-12}
  &&&& Tracking (reachable & Man. type & 4–6 RTN & \textless 1 orb. && \multirow4*{Reactive} & Check freq. $n_S$ & 10 s \\
  &&&& set + SOCP/QP-based)\cite{koenig_2021_opt} & Man. freq. & optimized & \textless 1–10 cm &&& Plane & 3D \\
  \cdashline{5-8}
  &&&& Tracking & Man. type & 2–3 RTN & \textless 1 orb. &&& Min. sep. $\epsilon$ & 5 m \\
  &&&& (least squares-based) & Man. freq. & time-fixed & \textless 1–10 cm &&& Confidence $q$ & 3 $\sigma$ \\
  \cdashline{5-8} \cline{10-12}
  &&&& Maneuver splitting & Max $\Delta \boldsymbol{v}$ bit & 2 mm/s & – \\
  &&&& (analytical + LP-based) & Man. split & 45 s & \\
  \cline{5-8}
\end{tabular}}
\end{table*}

\begin{figure}[ht!]
    \centering
    \includegraphics[width=0.75\linewidth]{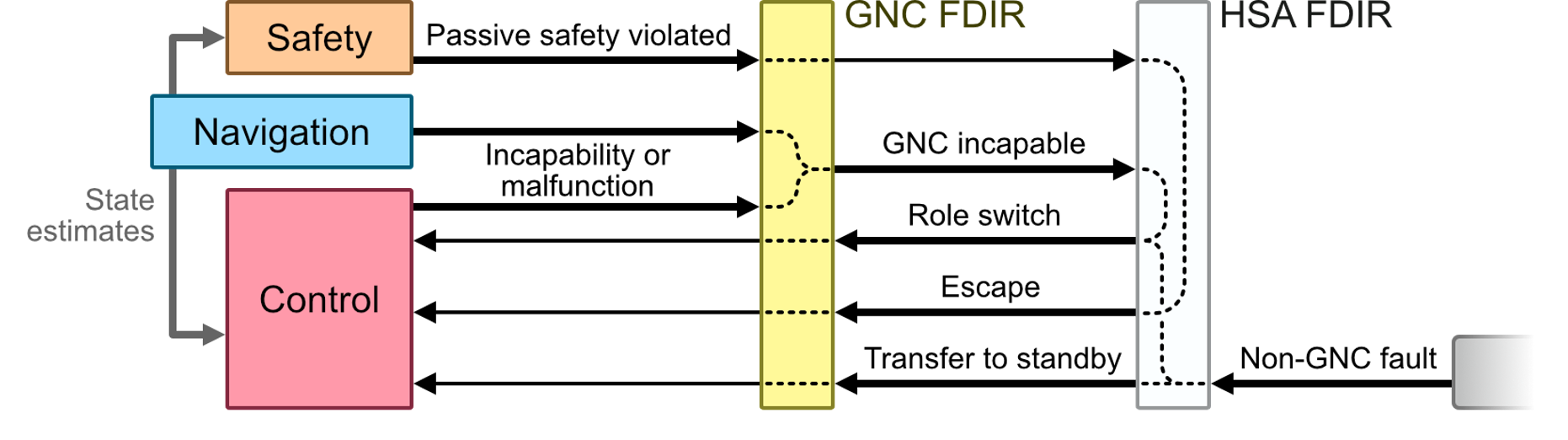}
    \caption{VISORS GNC FDIR}
    \label{fig:gnc-fdir}
\end{figure}

\paragraph{Fault detection, isolation and recovery (FDIR)} GNC interfaces with the HSA to perform FDIR (Figure \ref{fig:gnc-fdir}). Both GNC and non-GNC faults pass through the HSA FDIR block, which selects a response. Non-GNC faults trigger either a role switch between active (maneuvering) and passive (non-maneuvering) spacecraft, or a transfer to standby if the formation is in science or transfer. GNC faults trigger a role switch. Passive safety violations detected by the GNC safety module trigger an escape. The safety module receives state estimates from navigation and propagates them over a horizon $T$ to check for control-free separation violations at $q$-$\sigma$ confidence in RN or 3D (configurable). In case of escape, the control module plans escape maneuvers.

By integrating tightly with the HSA FDIR, all actionable faults can be handled by either a role switch, transfer to standby, or escape. For example, if the currently active spacecraft enters a bus safe mode, a role switch allows the other spacecraft to take control and continue nominal operations or activate a contingency plan. However, the mission remains single-fault intolerant. For example, a prolonged uncontrolled safe mode simultaneously on both spacecraft presents an irreducible collision risk accepted by the mission team.

\paragraph{On-board models} On-board models represent the on-board belief of: (i) environmental forces acting on the spacecraft (Table \ref{tab:dynamics}), (ii) spacecraft ballistic properties (Table \ref{tab:dynamics}), (iii) spacecraft internal geometry (Table \ref{tab:visors_int_comp}), and (iv) actuation behavior (Table \ref{tab:visors_act}). These models are used in navigation, control, and collision detection, and contain parameters affected by uncertainty. Model parameters can be updated from the ground, or internally if a prediction model is available or they are estimated within the navigation filter. Section 5 analyzes the sensitivity of GNC performance to key uncertain parameters. The remaining sources of uncertainty relate to GPS measurements and accuracy of the attitude solution from ADC through Time at Tone.

\begin{table*}[ht!]
\centering
\caption{Spacecraft ballistic features and dynamics models}\label{tab:dynamics}
\vspace{1mm}
\resizebox{\linewidth}{!}{
\begin{tabular}{ c c c c c c c}
 \hline
 \hline
  \multicolumn{2}{c}{Spacecraft features} & & \multicolumn{4}{c}{Dynamics models} \\
  \cline{1-2} \cline{4-7}
  Wet mass (kg) & \begin{tabular}{c c} DSC & 11.35 \\ OSC & 10.62 \end{tabular} & & & Ground-truth numerical \cite{giralo_phd_2021, visors2021koenig} & On-board numerical \cite{visors2021koenig} & On-board analytical \cite{koenig_2017_new, guffanti_long-term_2017} \\
  \cline{5-7}
  Cross section area (m\textsuperscript{2}) & Interpolated from CAD & &  Type &  GVE integration &  GVE integration &  State transition matrix \\
  & min-max: 0.15-0.55 & & Geopotential & GGM05S ($60 \times 60$) \cite{GRACE} & GGM01S ($20 \times 20$) \cite{GRACE} & $J_2$ secular \\
  $C_d$ (-) & 2.2 & & Atmospheric drag & NRLMSISE00 \cite{picone2002nrlmsise} & Harris-Priester & -- \\
  $C_r$ (-) & 1.8 & & Solar radiation pressure & \begin{tabular}{c} Analytical Sun emphemeris \\ Discrete cylindrical shadow \end{tabular} & \begin{tabular}{c} Analytical Sun emphemeris \\ Discrete cylindrical shadow \end{tabular} & -- \\
  \cline{1-2}
  & & & Third-body gravity & Analytical lunisolar ephemeris & Analytical lunisolar ephemeris & -- \\
  & & & Process noise & -- & 1 m/orbit & 1 m/orbit \\
  & & & Integrator & RK4 & RK4 & Euler \\
  & & & Step size & Dynamic 1-10 s & Dynamic 5–30 s & 5–30 s \\
  & & & State representation & \begin{tabular}{c} Quasi-nonsingular \\ orbit elements \end{tabular} & \begin{tabular}{c} Quasi-nonsingular \\ orbit elements \end{tabular} & Relative orbit elements \\
  & & & Usage & Ground-truth propagation & \begin{tabular}{c} Navigation time update, \\ control \& safety mean propagation \end{tabular} & \begin{tabular}{c} Maneuver solvers dynamics models, \\ control \& safety covariance propagation \end{tabular} \\
  \cline{4-7}
\end{tabular}}
\end{table*}

\subsection{4.2. Precise formation navigation}

\begin{figure}[h]
	\centering
    \includegraphics[width=\textwidth]{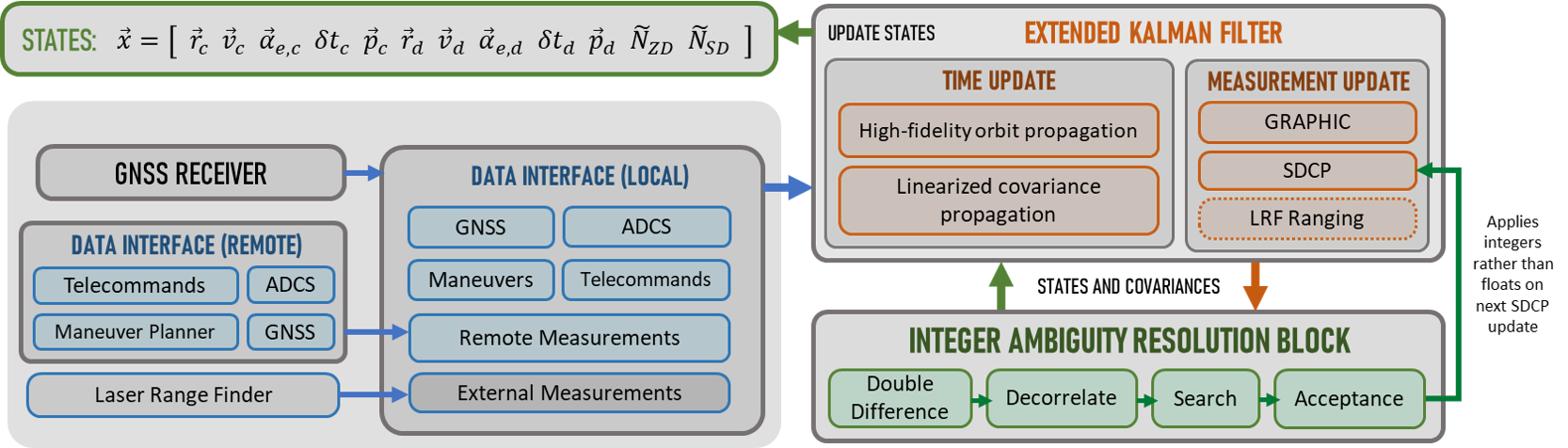}
    \caption{Navigation block diagram}
    \label{fig:dgtl-block-diag}
\end{figure}

Precise relative satellite navigation is achieved using a tailored variant of SLAB's Distributed Multi-GNSS Timing and Localization (DiGiTaL) flight software \cite{giralo2019digital, giralo2021digital}. DiGiTaL leverages the powerful error-cancellations of the Group and Phase Ionospheric Calibration (GRAPHIC) measurement \cite{yunck1993graphic} for absolute position estimation, and the Single-Difference Carrier Phase (SDCP) measurements for precise baseline estimation between the OSC and DSC \cite{damico_phd_2010}. This is followed by real-time, on-board integer ambiguity resolution (IAR) of double-differenced CDGPS ambiguities. The major blocks of the DiGiTaL flight software, depicted in Figure \ref{fig:dgtl-block-diag}, are: (1) the data interface block which handles messages, validation of measurement time-tags, coordinate transforms etc; (2) the orbit determination block, which employs an efficient Extended Kalman Filter (EKF); and (3) the integer ambiguity resolution block, which performs integer decorrelation, search, and resolution on double-differenced carrier phase integers using the Modified Least Squares Ambiguity Decorrelation Adjustment algorithm, or mLAMBDA \cite{teunissen1994lambda, chang2005mlambda}. The navigation filter of each spacecraft tracks the states

\begin{equation}
\label{eq:states}
    \Vec{X} = \left[ 
    \Vec{r}_c, \ \Vec{v}_c, \ 
    \Vec{\alpha}_{e,c}, \ c \delta t_c, \Vec{p}_c,
    \Vec{r}_d, \ \Vec{v}_d, \ 
    \Vec{\alpha}_{e,d}, \ c \delta t_d, \Vec{p}_d,
    \Tilde{N}_{ZD}, \Tilde{N}_{SD} \right]
\end{equation}

\begin{figure}[h]
	\centering
    \includegraphics[width=\textwidth]{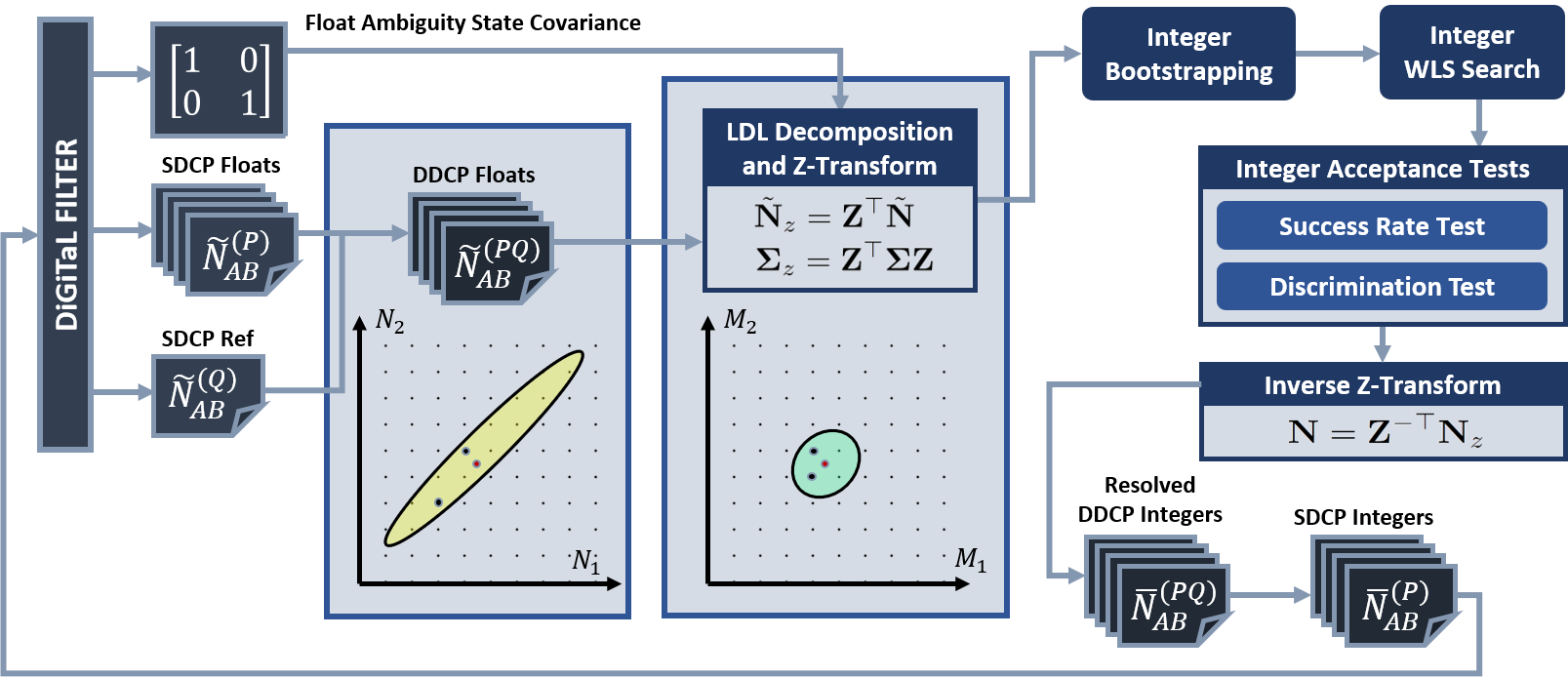}
    \caption{Integer ambiguity resolution block diagram}
    \label{fig:dgtl-iar-block}
\end{figure}

where $\Vec{r}_c$, $\Vec{v}_c$, $\Vec{r}_d$, $\Vec{v}_d$, are the Earth-centered inertial (ECI) positions and velocities of the chief (DSC) and deputy (OSC) respectively. Dynamic model compensation through the estimation and application of stochastic empirical acceleration terms $\Vec{\alpha}_{e,c}$, $\Vec{\alpha}_{e,d}$ account for unmodelled dynamics \cite{stacey2021adaptive}. These are applied as perturbative forces in the radial-tangential-normal (RTN) reference frame of the DSC, and added to the force model of the orbit propagation step in the filter time update. Several measurement and kinematic biases are also estimated, such as the receiver clock offsets $c \delta t_c$ and $c \delta t_d$ and the body-frame antenna phase center offsets with respect to the static center of mass $\Vec{p}_c$ and $\Vec{p}_d$ as illustrated in Figure \ref{fig:visors-int-comp}. Specific to carrier phase biases, the undifferenced ambiguities of the local spacecraft $\Tilde{N}_{ZD}$ and the SDCP ambiguities $\Tilde{N}_{SD}$ between the remote and local spacecraft are estimated as floats, and resolved into integers after a successful discrete search with acceptance tests \cite{teunissen1998success}, as depicted in Figure \ref{fig:dgtl-iar-block}. For algorithmic details of the IAR process in DiGiTaL, the reader is invited to peruse references by Giralo et al \cite{giralo2019digital, giralo2021digital}. The maximum number of tracked ambiguities is 24 each for $\Tilde{N}_{ZD}$ and $\Tilde{N}_{SD}$. In total, the state vector $\Vec{X}$ comprises 74 elements of floats.

The critical innovation that enables DiGiTaL's practicality is its ability to balance computational timeliness with navigation precision in the baseline estimation while running both the filtering and IAR block. The size of the state vector results in a $74 \times 74$ covariance matrix which can be computationally challenging for an on-board CubeSat-grade flight computer to perform operations with. As such, multiple computational optimizations were undertaken. The Extended Kalman Filter was adopted over the Unscented Kalman Filter in the original DiGiTaL package \cite{giralo2021digital} without compromising accuracy at small separations. The state mean vector and covariance matrix are dynamically re-sized (smaller) when possible. Sparse matrix operations are applied where possible. Symmetry in matrices during IAR are exploited \cite{chang2005mlambda}. Both the DSC and OSC run an instance of DiGiTaL on-board, independently estimating the states in Equation \ref{eq:states}, and relying on the cross-link to exchange measurements, maneuvers, attitude and time with the remote spacecraft.

\subsection{4.3. Precise formation control}

\begin{figure}[ht!]
    \centering
    \includegraphics[width=0.9\linewidth]{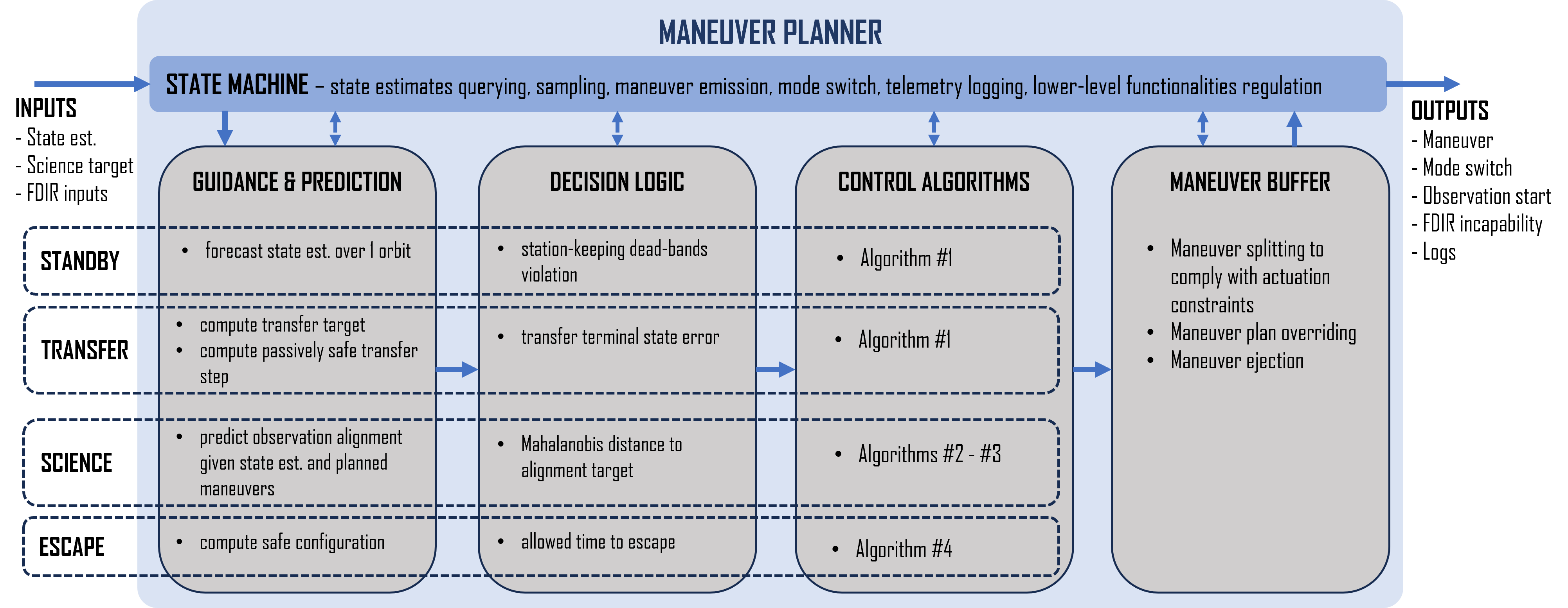}
    \vspace{1mm}
    \caption{Control stack, with components specialized per mission mode}
    \label{fig:gnc-ctrl-stack}
\end{figure}

\paragraph{Control stack} The control component, or maneuver planner, uses a consistent stack architecture across mission modes (Figure \ref{fig:gnc-ctrl-stack}), with algorithms within each layer specialized from mode to mode. Control acts upon three inputs: 1) state estimates from navigation, 2) ground commands to start/end a science campaign, and 3) FDIR responses (i.e., escape trigger, transfer to standby, or enabling/disabling control due to a role switch). During mission modes that prescribe a certain relative orbit (i.e. standby and science), control propagates the current state estimate to the control horizon, decides whether the current maneuver plan needs to be updated, and computes new maneuvers if so. Transfer mode additionally: (i) computes the transfer target (e.g., science relative orbit for observation alignment, or the standby orbit), and (ii) computes intermediate guidance waypoints to enable passively safe transfer between initial and final relative orbits. At any time, upon command from HSA FDIR, the maneuver planner can plan one or more escape maneuvers that enforce safe separation in the RN-plane and slow repulsive drift in the tangential direction. Finally, the maneuver planner includes a buffer to spread and store planned maneuvers without commanding them to the HSA until execution. This meets maneuver quantization constraints imposed by actuation (Table \ref{tab:visors_act}), and allows altering previously computed maneuvers for closed-loop corrections or contingecy scenarios.

\paragraph{Relative orbit control algorithms} Even if the algorithms used for relative orbit control change by mode, allowing for tracking/station-keeping, reconfiguration and escape, the underlying optimal control problem to be solved is ultimately the same. In particular, it can be formalized mathematically as a fuel-optimal (minimum $\Delta \boldsymbol{v}$) two-point-boundary-value-problem:
\begin{equation}
\label{OCP_IC}
\begin{aligned}
& \underset{\Delta \boldsymbol{v}(t)}{\text{minimize}} & & \int_{t_e}^{t_f} || \Delta \boldsymbol{v}(t) ||_2  dt &\\
& \text{subject to} & & \Delta \delta \textbf{\oe} = \delta \textbf{\oe}(t_f) - \boldsymbol{\Phi}(t_e, t_f) \delta \textbf{\oe}(t_e) = \int_{t_e}^{t_f} \boldsymbol{\Gamma}(t) \Delta \boldsymbol{v}(t) dt \\
&&& \delta \textbf{\oe}(t_f) \equiv \textrm{guidance} & \\
&&& \delta \textbf{\oe}(t_e) \equiv \textrm{state estimate} & \\
\end{aligned}
\end{equation}
where $\delta \textbf{\oe} \in \mathbb{R}^6$ is the quasi-non singular relative orbit elements (ROE) state \cite{damico_phd_2010} of the instantaneously active spacecraft with respect to the passive spacecraft. $\Delta \boldsymbol{v}(t) \in \mathbb{R}^3$ is delta-velocity applied by the active spacecraft expressed in the RTN frame centered in the passive spacecraft. The initial condition is the most updated state estimate provided by the navigation filer. The target condition is provided by the on-board guidance and represents either a relative orbit correction (for tracking during standby and science), or a reconfiguration way-point (for transfer), or a safe target configuration (for escape). The affine dynamics constraint includes the analytical $J_2$ perturbed model presented in Table \ref{tab:dynamics} on the right. To minimize fuel consumption the minimum 2-norm of the applied $\Delta \boldsymbol{v}$ is sought. As described in Table \ref{tab:gnc-algos} and Figure \ref{fig:gnc-ctrl-stack}, depending on the mission mode, this optimal control problem is solved in four different ways allowing for a desirable range of terminal control accuracies (from meter-level, down to centimeter-level), varying control hozions (from few orbits, down to fraction of an orbit), and maneuver frequency and timeliness. The four implemented control algorithms are:

\begin{itemize}[topsep=0pt] 
    \item[$\triangleright$] \textit{Algorithm \#1} : \textbf{Closed-form time-optimized algorithm} -- This control algorithm computes the optimal maneuver location and magnitude analytically in closed-form \cite{gaias_imp_2015, chernick_2018_closed}. In particular, triplets of in-plane maneuvers and one/two out-of-plane maneuvers are implemented every multiple of half-an-orbit to achieve meter-level control accuracy over horizons greater or equal to 1.5 orbits. This algorithm is used for standby mode relative orbit station-keeping, to transfer between intermediate passively-safe way-points during transfer mode, for initial relative orbit acquisition during commissioning, and for standby orbit reaquisition after an escape maneuver. This algorithm has partial flight heritage in the PRISMA \cite{gill_autonomous_2007} and BIROS \cite{gaias_avanti_2017} missions.
    \item[$\triangleright$] \textit{Algorithm \#2} : \textbf{Convex optimization and reachable set theory-based time-optimized algorithm} -- This control algorithm leverages reachable-set theory to compute both the optimal maneuver location and magnitude by solving two convex optimization problems \cite{koenig_2021_opt}. A Second Order Cone Program (SOCP) is formalized to solve for optimal maneuver candidate times, whereas a Quadratic program (QP) is formalized to solve for corresponding optimal maneuver magnitudes. The main advantages of using this optimization-based algorithm are  that: (i) the optimal maneuver times are solved for directly through reachable set theory, without the need of introducing a dense time discretization of the control window (and associated optimization variables) as commonly done in direct-optimization-based control approaches \cite{Betts98}, (ii) that an optimal maneuver sequence of usually $4$-$6$ maneuvers can be find over control horizons down to a fraction of an orbit. During the mission this allows to fine track the science relative orbit leading to observations, with a progressively shrinking control horizons. To achieve centimeter-level accuracy, the algorithm is used in closed-loop within a stochastic model predictive control (MPC) pipeline. Specifically, at each state estimate acquisition the relative motion is propagated in open-loop up to observation accounting for the previously computed control plan. If the Mahalanobis distance to target is below a threshold the plan is kept, otherwise is recomputed given the latest state estimate. From an implementation stand-point, the convex optimization problems are solved on-board using the Embedded COnic Solver (ECOS) \cite{ecos_13}, an interior-point solver for SOCPs \cite{boyd_book}. The technology demonstration in-space of such a convex-optimization-based translational control approach is a novelty for CubeSats and a contribution brought by VISORS.
    \item[$\triangleright$] \textit{Algorithm \#3} : \textbf{Closed-form Tikhonov-regularized least-square-based time-fixed algorithm} -- During the very terminal phases prior to scientific observation (10–15 min before), it is effective to compute a 2–3 impulse maneuver sequence at fixed times that complies with attitude slewing and settling requirements (usually, as soon as possible, $\sim$45 s before observation to allow for attitude settling, and in the middle). This is achieved by solving on-board a Tikhonov-regularized least squares problem, which directly solves for maneuver $\Delta \boldsymbol{v}$ components at the specified time instants while penalizing fuel consumption. In particular, defining with $\Delta \bar{\boldsymbol{v}} \in \mathbb{R}^{3 \times N}$ the stacked control input sequence over the fixed maneuvering times $t_1, \dots t_N$ ($N = 2, 3$), the following optimization problem
    \begin{equation}
    \Delta \bar{\boldsymbol{v}}^* = \arg\min_{\Delta \bar{\boldsymbol{v}}} {|| \boldsymbol{\Gamma} \Delta \bar{\boldsymbol{v}} - \Delta \delta \textbf{\oe} ||^2 + \nu || \Delta \bar{\boldsymbol{v}} ||^2 }
    \end{equation} 
    can be solved in closed form as 
    \begin{equation}
    \Delta \bar{\boldsymbol{v}}^* = \left( \boldsymbol{\Gamma}^T \boldsymbol{\Gamma} + \nu \boldsymbol{I} \right) \boldsymbol{\Gamma}^T \Delta \delta \textbf{\oe}
    \end{equation}
    \item[$\triangleright$] \textit{Algorithm \#4} : \textbf{Escape algorithm} -- In the scenario where passive safety is predicted to be violated, a quickly executable escape maneuver is computed to enforce a minimum RN separation between spacecraft while increasing the along track separation. The escape maneuver is computed as a single impulse that achieves a minimum RN separation by analytical condtions on the ROE\cite{koenig_2018_robust} and along track drift rate prescribed by a tuned relative semi-major axis. This is inherently a guidance problem to find an ROE state which can satisfy the safety conditions and be reached with a single impulse at an arbitrary time, which can be solved exactly in closed form via the Gauss Variational Equations for the ROE \cite{damico_phd_2010}. The time of the maneuver is selected to as soon as possible after the contingency is detected while permitting time for the execution of buffered maneuvers. Along track drift is driven by the relative semimajor axis; to ensure the desired direction and minimum drift rate in the presence of uncertainty in the estimated state, the $\delta a$ variation is compensated to reach its nominal value with $3$-$\sigma$ error.
\end{itemize}

\paragraph{Passively safe transfer guidance}
One of the most challenging phases from a fault-tolerant safety perspective is the transfer between standby and science relative orbits. Both standby and science modes are designed in closed-form to achieve passive safety in the RN-plane through relative eccentricity/inclination-separation \cite{damico_proximity_2006}. Nevertheless, passive safety has not only to be guaranteed at initial and terminal states, but at every instant of the transfer too. In particular, during transfer, the radial-normal separation shrinks from 200 m (standby) to 20 m (science). Multiple solutions exist to ensure passive safety along the transfer. A computationally efficient option is to realize the transfer stepwise, moving through subsequent passively safe waypoints using a closed-form two-point boundary value problem solver as the one described in the previous section \cite{gaias_imp_2015, chernick_2018_closed}. This approach does not allow for end-to-end transfer optimization and does not provide hard guarantees of passive safety on the trajectory in between waypoints. Therefore, it must rely on fallback safety strategies, such as the implemented collision detection and escape logic, which checks for passive safety violations every time a new state estimate is available, and triggers an escape upon detection. Nevertheless, the chances of passive safety violation can be reasonably minimized by clevery selecting the transfer waypoints. In particular, the transfer can be implemented as a sequence of 1.5-orbit reconfiguration steps, designed such that: (i) the total $\Delta \boldsymbol{v}$ of each step scales proportionally to the spacecraft RN separation, (ii) the relative eccentricity and inclination vectors move synchronously while remaining almost parallel or anti-parallel, (iii) the relative semi-major axis variations scale proportionally to the spacecrat RN separation to avoid compromising the passive safety margin (by translating the RN ellipse in radial direction too much). Note that the relative eccentricity/inclination vectors in science configuration are affected by the reference orbit beta angle (see Section 3), as well as by the selection of the pole at which observations happen. Pathological situations that cause possible passive safety violation during the transfer are when the computed eccentricity/inclination vectors in science are antiparallel whereas the standby ones are parallel, or vice versa. Such situations are avoidable by properly selecting the standby ROE when the reference orbit beta angle is fixed upon launch. Testing results of this stepwise passively safe transfer approach are presented in Section 5.

\paragraph{Maneuver buffering} In order to satisfy the actuator constraints on the magnitude and frequency of impulses, a methodology has been introduced for executing a large maneuver as the net sum of a series of constraint abiding maneuvers. A maneuver $\Delta \boldsymbol{v} = [\Delta v_R \ \Delta v_T \ \Delta v_N]$ occuring at mean argument of latitude $u$ is decomposed into a series of identical maneuvers spaced $\Delta u$ apart from the original maneuver location. The extended maneuver plan realizes the same net effect on the ROE as the orignal planned maneuver using $2N+1$ maneuvers $\Delta \boldsymbol{v}_i $ in RTN and $2M$ maneuvers $\Delta \boldsymbol{v}_j $ in the R direction. Following Algorithm 5 below, the extended maneuver components can be calculated by iteratively introducing maneuvers until the magnitude constraint is satisfied. 

\begin{algorithm}[h]\footnotesize
\caption{Maneuver buffering}\label{ALG_buffer}
   \DontPrintSemicolon
   
   \KwIn{$\Delta \boldsymbol{v}$, $u$}
   
   \KwOut{$\Delta v_{Ri}$, $\Delta v_{Ti}$, $\Delta v_{Ni}$, $\Delta v_{Rj}$, N, M }
   
   \KwData{$\Delta v_{max}$, $\Delta u$}
   
   \Begin{ 
   N, M = 0 \;
   \While{$\|[\Delta v_{Ri} \ \Delta v_{Ti} \ 
   \Delta v_{Ni}]\|_1 > \Delta v_{max}$}{
      N+=1 \;
   $\Delta v_{Ri} =  \Delta v_{R}\left( 1 + 2 \sum_{i=1}^{N}\cos(\Delta u * i) \right)^{-1}$ \;
   $\Delta v_{Ti} =  \Delta v_{T}\left(1+2N\right)^{-1}$ \;
   $\Delta v_{Ni} =  \Delta v_{N}\left( 1 + 2 \sum_{i=1}^{N}\cos(\Delta u * i) \right)^{-1}$ \;
   
   }
   \While{$|\Delta v_{Rj}| > \Delta v_{max}$}{
   M+=1 \;
   $\Delta v_{Rj} =  \left(\Delta v_{T} - \Delta v_{Ti} *(1+2N) \right)* \left(2\sum_{i=1}^{M}\sin(\Delta u *(N+ i)) \right)^{-1}$ \;
   }
   \Return  \;
   }
\end{algorithm}


\section{5. Software-in-the-loop testing and validation}
\label{section5}

Simulation testing for VISORS GNC has two objectives. First, to assess the system-level behavior and performance of the GNC system during nominal and off-nominal mission modes while interfacing with other subsystems and ground-truth orbit dynamics, and second, to analyze the sensitivity of observation alignment performance to key uncertain system parameters via parameter sweeps and Monte Carlo simulation. Within simulations, ground-truth dynamics are modeled as in Table \ref{tab:dynamics}, and key subsystems are simulated to mimic real on-board behavior (e.g., propulsion as in Table \ref{tab:visors_act}).

\begin{figure}[ht!]
    \centering
    \includegraphics[width=0.9\linewidth]{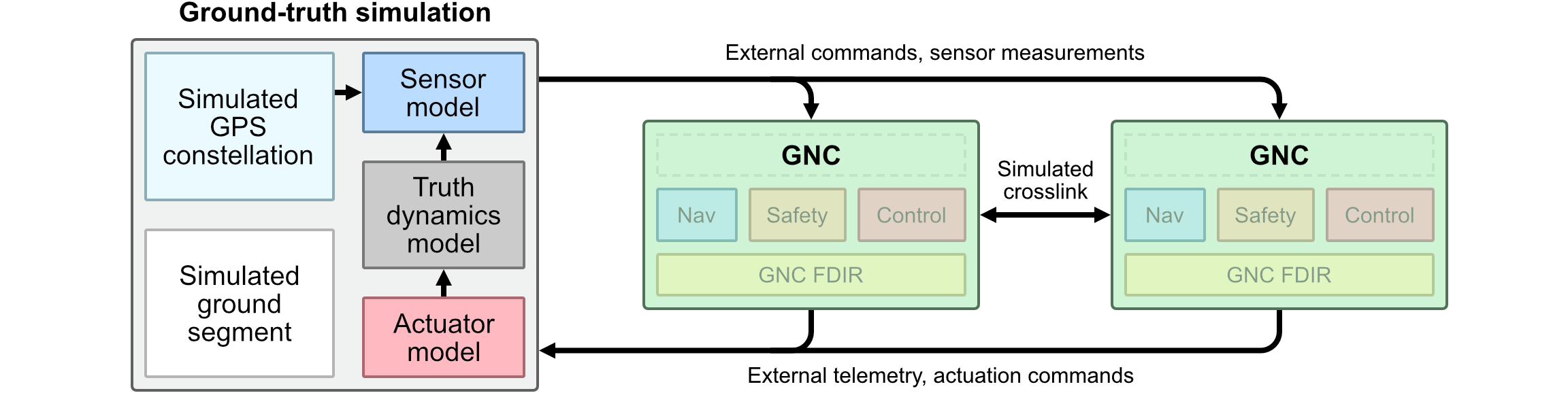}
    \caption{GNC software-in-the-loop simulation architecture}\label{fig:gnc-testing}
\end{figure}

The first objective—system-level performance and interface testing—is accomplished via a custom software-in-the-loop simulation architecture (Figure \ref{fig:gnc-testing}) capable of exercising the GNC system interface with high fidelity. To increase the chance that performance in simulation accurately predicts performance in flight, the simulation models a wide range of realistic effects: high-fidelity orbit perturbations, timing synchronization caused by crosslink transmission delays, spacecraft attitude impacts on drag area and visible GPS satellites, variations in GPS satellite orbits, anticipated solar activity in October 2024, etc. The high fidelity of the simulation software has been a critical asset for developing the VISORS GNC system.

The second objective—analyzing GNC's sensitivity to uncertain parameters—is accomplished via two forms of simulation parameter variation: first, explicit parameter sweeps in which simulations are run varying a specific parameter while holding all else constant; second, Monte Carlo simulations in which many ground-truth parameters are varied around a mean according to a specified distribution. Observation alignment performance as a function of error in the center of mass position knowledge (Table \ref{tab:visors_int_comp}) and maneuver execution error in direction and magnitude (Table \ref{tab:visors_act}) is shown in Figure \ref{MC_align_metr}.

\begin{table}[ht!] 
\centering
\caption{VISORS mission test case scenario}\label{tab:visors_test}
\resizebox{0.8\linewidth}{!}{
\begin{tabular}{ c c | c c c | c c }
 \hline 
 \hline
  \multicolumn{2}{c}{Reference orbit} & \multicolumn{3}{c}{Relative orbit modes} & \multicolumn{2}{c}{Campaign \& Safety} \\
  \cline{1-2} \cline{3-5} \cline{6-7}
   Element & SSO & Element & Standby (m) & Science (m) & \\
   \cline{1-7}
   $a$ (km) & $R_E$ + 500 & $a \delta a$ & 0 & –2.62 & observations (\#) & 10 \\
   $e$ (-) & 0.004 & $a \delta \lambda$ & 0 & 45.21 & PS type & RN \\
   $i$ (deg) & 97.8 & $a \delta e_x$ & 0 & –34.51 & $T$ (h) & 1.5 \\
   $\Omega$ (deg) & 157.5 & $a \delta e_y$ & 200 & 4.78 & $\epsilon$ (m) & 5 \\
   $u(t_0)$ (deg) & 0 & $a \delta i_x$ & 0 & –18.72 & $q$ ($\sigma$) & 3 \\
   LTAN & 10:00AM & $a \delta i_y$ & 200 & 2.72 \\
   Period (hr) & 1.58 & & & & \\
   $t_0$ (GPS) & 12 AM, 1 Oct 2024 & & & & \\
  \hline
  \hline
\end{tabular}}
\end{table}

\subsection{Navigation filter performance}

The integrated test scenario abides by the conditions prescribed in Table \ref{tab:visors_test}. The scenario begins with the DSC/OSC in standby mode for 45 minutes, before transferring to science, performing 10 observation attemps, and transferring back to standby. This test campaign lasts 35.5 orbits ($\sim$ 2.3 d). The nominal attitude profile has the GPS antenna boresight maintained in zenith-pointing, with an assumed elevation mask of \textgreater 5$^\circ$ from the local horizontal of the antenna plane. An attitude slew maneuver is executed at a maximum slew rate of 1.5$^\circ$/s when transiting between zenith pointing attitude into science mode observations. 

The dynamics follows Table \ref{tab:dynamics}: the ground truth trajectory applies the ground-truth numerical integration; the navigation filter applies the on-board analytical model for the state covariance propagation, and the on-board numerical integration for the filter mean state propagation. GPS ephemeris errors are injected as a corruption of GPS vehicle orbital elements in the form of ROEs into the GPS broadcast ephemeris message. This results in introducing periodic variations of the Cartesian error, rather than random errors, in both position and velocity of the GPS satellite, as predicted by the linear mapping between ROE and Cartesian space. Receiver and GPS satellite clock errors are modelled as a random walk with with step size of 1m applied each second. Ionospheric delays are applied via the Klobuchar model. Thermal noise of 20 cm and 2 mm are applied for undifferenced pseudorange and carrier phase respectivel. Center of mass variations of up to 3 mm are applied. A GPS antenna phase center offset of 8.54 cm along the boresight is applied from the antenna base, and boresight mounting errors of 30 arcsec are applied for all sensors as per Table \ref{tab:visors_int_comp}.

\begin{figure}[ht!]
    \centering
    \begin{subfigure}[ht!]{0.45\textwidth}
        \centering
        \includegraphics[width=\textwidth]{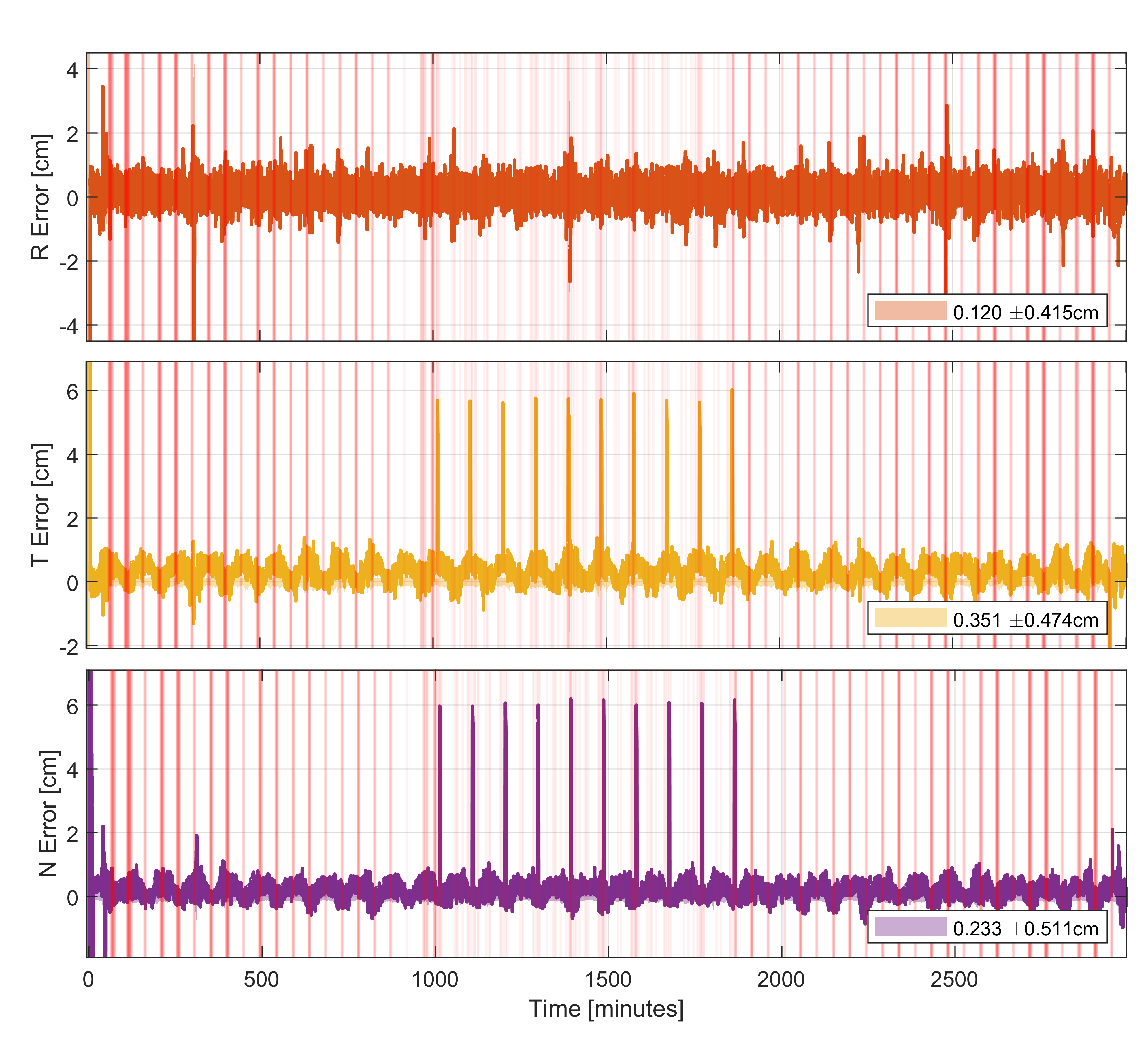}
    \end{subfigure}
    \hspace{0.3cm}
    \begin{subfigure}[ht!]{0.45\textwidth}
        \centering
        \includegraphics[width=\textwidth]{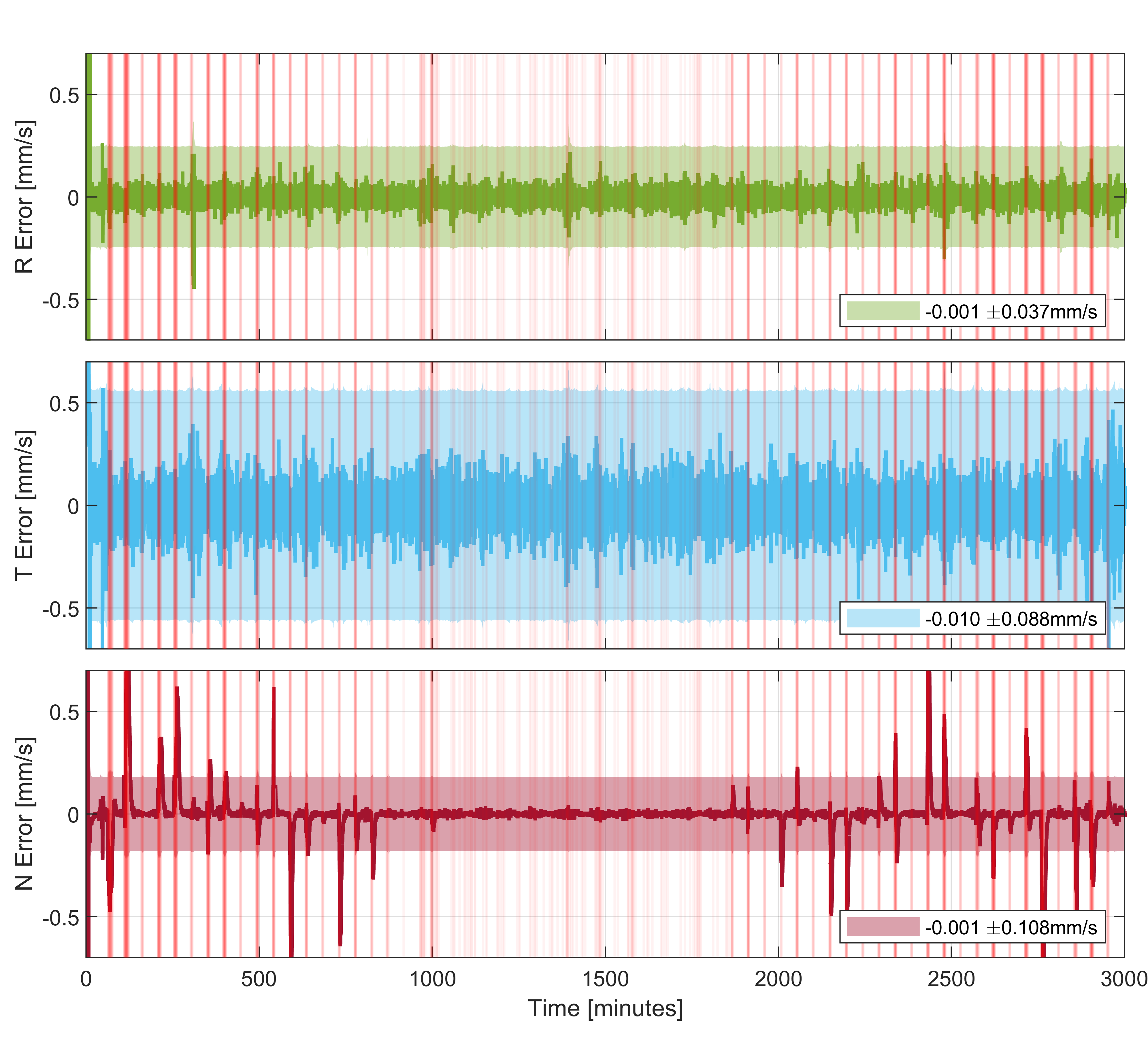}
    \end{subfigure}
    \caption{OSC-to-DSC relative position error (left); relative velocity error (right), as estimated by OSC}
    \label{fig:dgtl-results-1}
\end{figure}

\begin{figure}[ht!]
    \centering
    \begin{subfigure}[ht!]{0.45\textwidth}
        \centering
        \includegraphics[width=\textwidth]{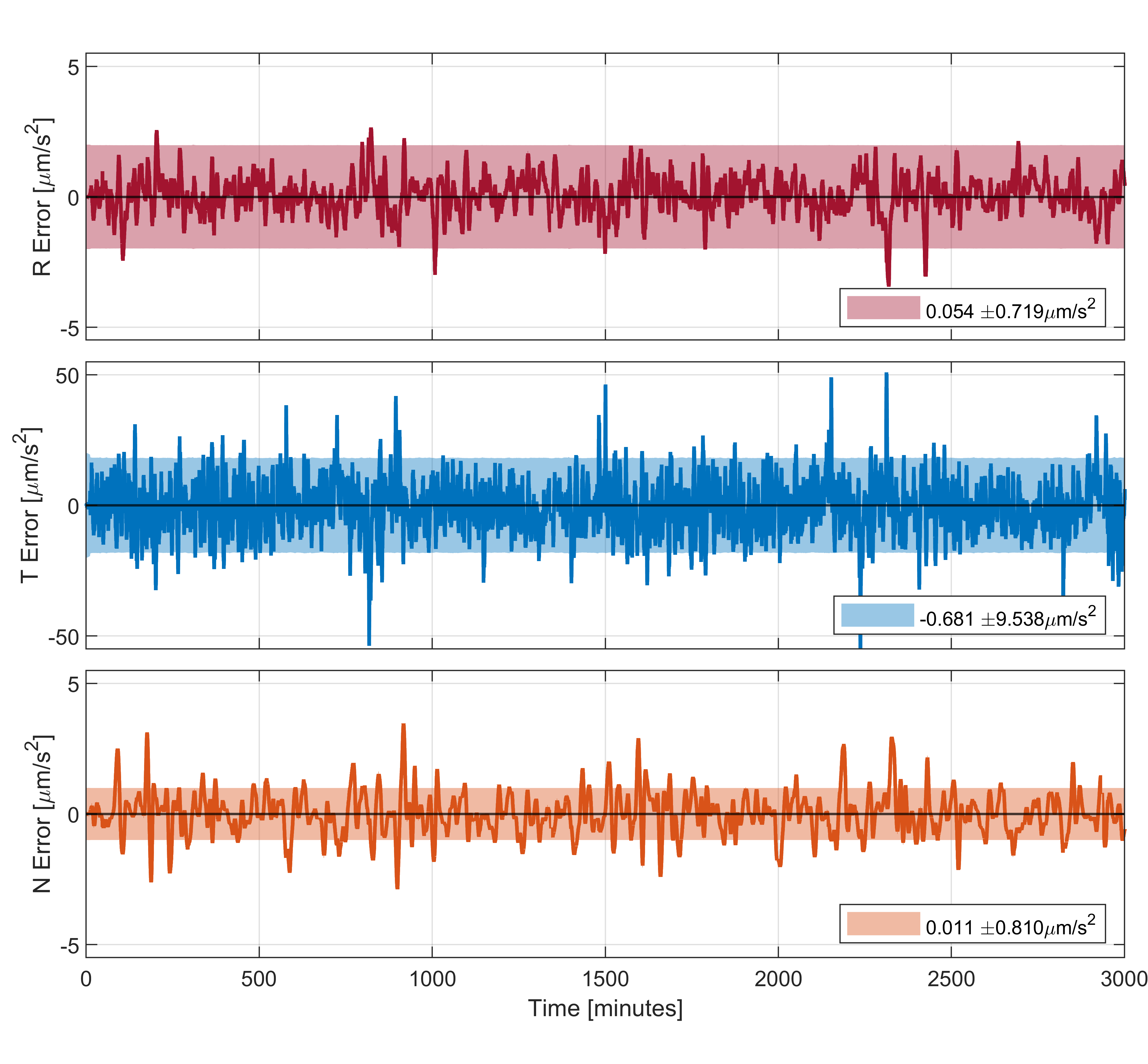}
    \end{subfigure}
    \hspace{0.3cm}
    \begin{subfigure}[ht!]{0.5\textwidth}
        \centering
        \includegraphics[width=\textwidth]{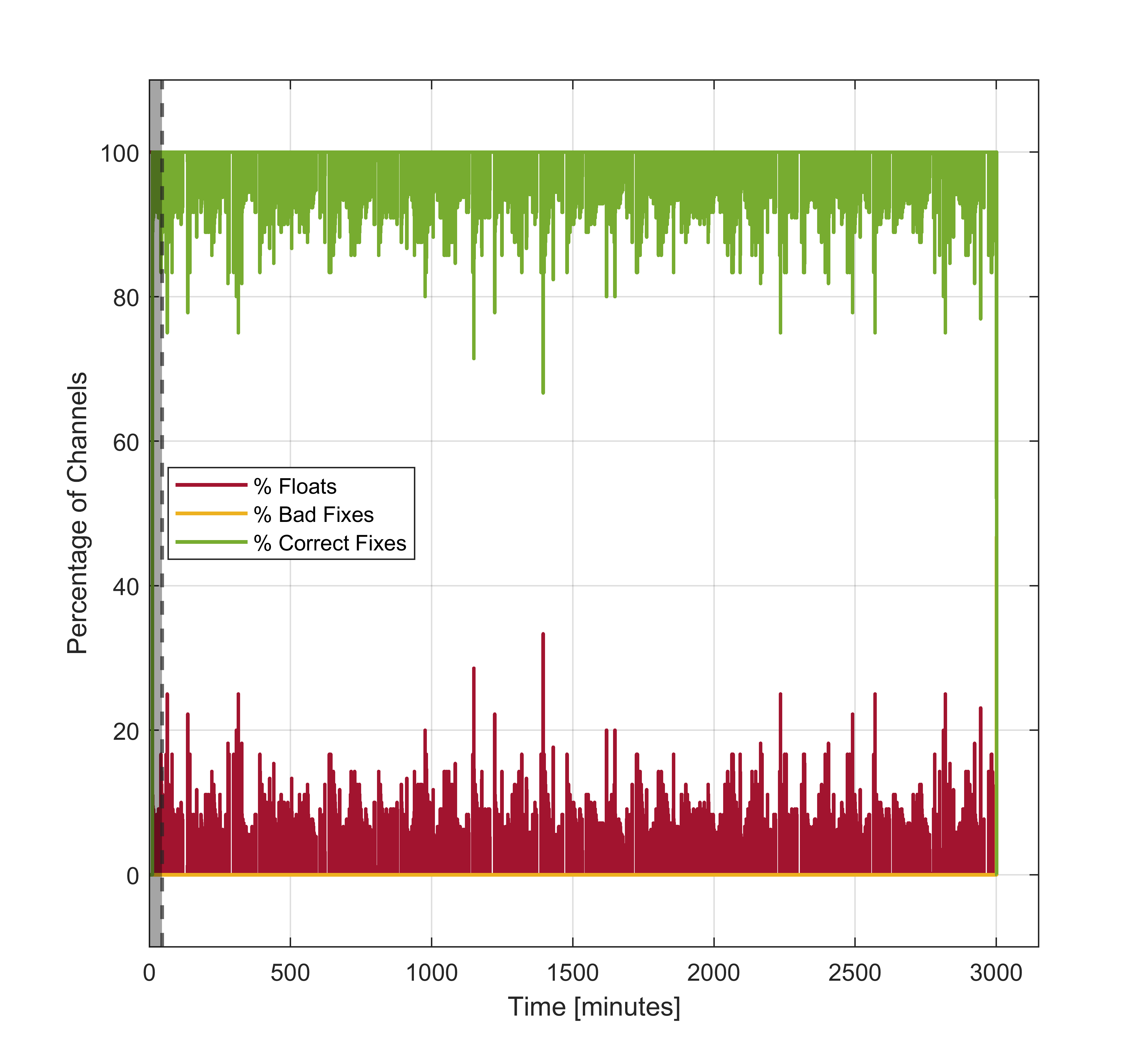}
    \end{subfigure}
    \caption{OSC estimated empirical accelerations in RTN (left) and IAR performance (right)}
    \label{fig:dgtl-results-2}
\end{figure}

Navigation performances of the OSC (active) with respect to the DSC (passive) are plotted in Figure \ref{fig:dgtl-results-1}. Executions of 2 mm$\cdot$s\textsuperscript{-1} impulsive maneuvers are illustrated by the translucent red bars. Maneuvers by the OSC are communicated to and accounted for by the DSC over crosslink with a simulated 0.5 s delay. Referring to Figure \ref{fig:dgtl-results-1}, there are 10 noticeable error spikes in relative state errors up to a magnitude of 8 cm RMS. These spikes coincide with the instances right before each of the 10 observation attempts. The primary reason for each brief spike before observation is due to the switching of attitude profiles from zenith-pointing (with sun-pointing constraints) to target observation. In doing so, one or more commonly observed GPS satellites may drop out of the elevation masks of both DSC and OSC. Commonly observed GPS satellites are necessary to form single-difference carrier phase measurements. The secondary reason is that observations take place at the poles, when the relative science orbit trajectory intersects with the TN-plane of the DSC (Figure \ref{fig:visors-observations}). This further degrades navigation accuracy because fewer GPS satellites are visible at the poles. Thus, the geometric dilution of precision increases shortly before observations. However, this does not affect the navigation performance during observation attempts, because IAR is maintained (Figure \ref{fig:dgtl-results-2}) and the filter is confident enough to fix any new ambiguities after completing attitude slewing.

Figure \ref{fig:dgtl-results-2} shows the OSC's filter's estimated empirical accelerations in which dynamic model compensation can be seen to successfully capture the effects of using different atmospheric drag models between flight and ground truth. In particular, the tangential axis (T) has a scale and standard deviation ten-fold that of the radial (R) and normal (N) axes, matching the differences in order of magnitude expected. As per Table \ref{tab:dynamics}, the ground truth employs the NRLMSISE-00 atmospheric density model with a predicted F10.7 Solar Flux on October 2024, which corresponds very closely to a solar maximum, while the flight model employs the Harris–Priester atmospheric density model in the filter using mean solar activity \cite{montenbruck_satellite_2012}. This differential effect reveals itself as much larger standard deviation in the tangential (T) axis acceleration.

\begin{figure}[ht!]
    \centering
    \includegraphics[width=0.75\linewidth]{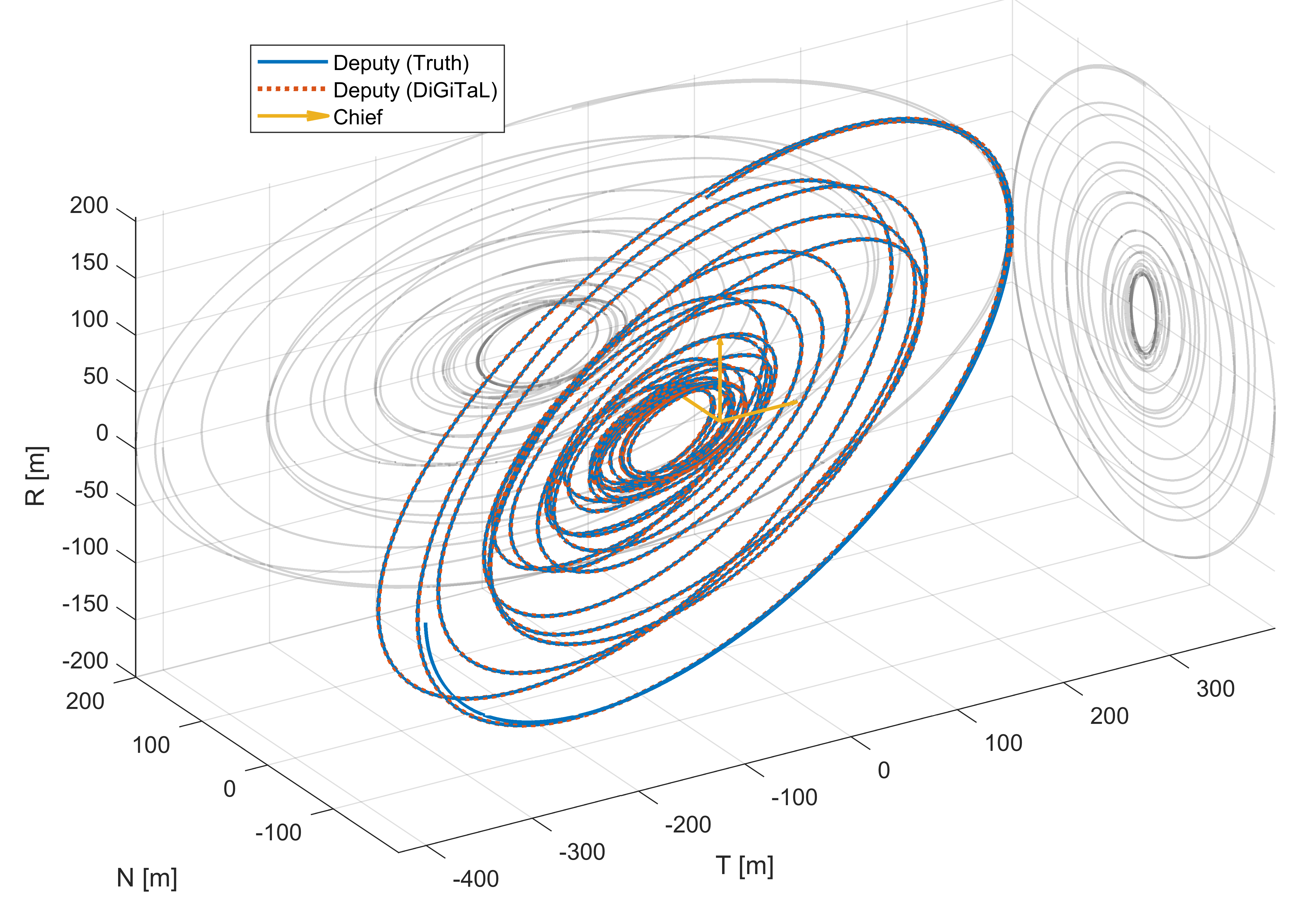}
    \caption{Estimated and ground-truth relative trajectory including standby, transfer to science, and transfer to standby (RTN frame, DSC chief). Projections onto the RT- (left) and RN- (right) planes shown in gray.}
    \label{fig:dgtl-trajectory}
\end{figure}

IAR is achieved successfully about 30 minutes into the campaign, achieving sub-cm and mm/s 1-$\sigma$ relative navigation confidence. The rate of successful integer fixes is 99.1171\% for the DSC and 99.0879\% for the OSC. Wrong fix rates are $\sim$ 0.008\% for both DSC and OSC (with remaining \% as unresolved float ambiguities). The campaign proved promising IAR performance (Figure \ref{fig:dgtl-results-2}) despite the time-varying baseline between DSC and OSC throughout the campaign. Figure \ref{fig:dgtl-trajectory} shows the relative trajectory of the campaign.

\subsection{Mission simulations: combined navigation and control performance testing}

For the scenario described in Table \ref{tab:visors_test}, this section presents the combined navigation and control results for the different mission modes. Appendix 7.A presents the associated delta-v budget.

\subsubsection{Standby station-keeping}
\begin{figure}[ht!]
    \centering
    \includegraphics[width=0.48\linewidth]{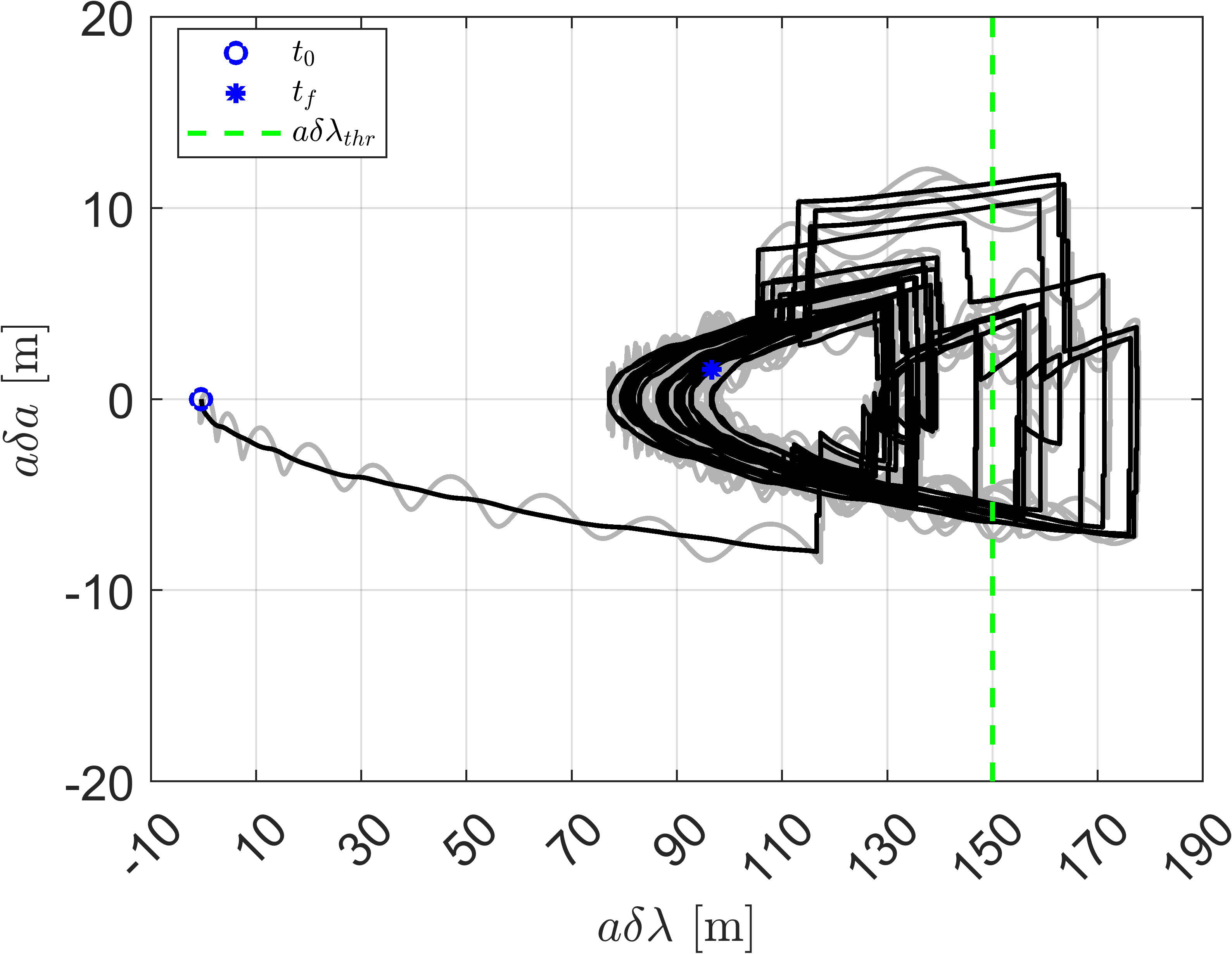} 
    \hspace{0.2cm} \includegraphics[width=0.46\linewidth]{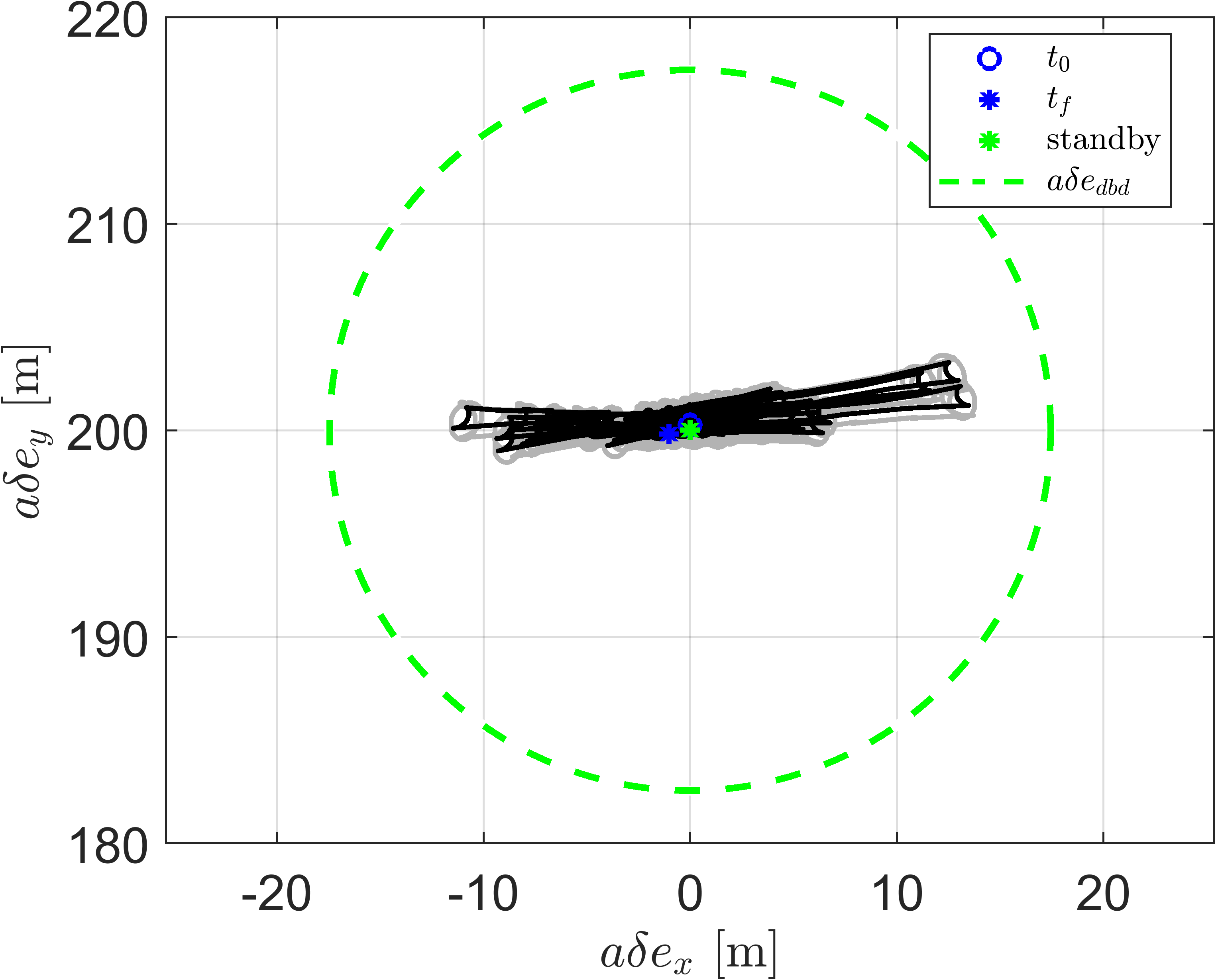} 
    \caption{Relative orbit elements (ROE) behavior during 1 week of standby relative orbit station-keeping}\label{ROE_standby}
\end{figure}

Figure \ref{ROE_standby} presents the evolution of the in-plane ROE during standby mode station-keeping. The black lines are the mean ROE, the grey lines are the osculating ROE. Station-keeping is largely in-plane dominated by the effect of $J_2$ and, in particular, differential drag. During standby mode, the maneuver planner state-machine down-samples the state estimates fed by navigation every second, and once per orbit propagates the relative orbital motion forward for one orbit. If either the relative eccentricity vector deadband $a \delta e_{dbd}$ or the threshold $a \delta \lambda_{\mathrm{thr}}$ (dashed green lines) are violated, a sequence of three tangential maneuvers is planned. This sequence is buffered and quantized in sub-impulses (using Algorithm \ref{ALG_buffer}) to comply to the actuation constraints in Table \ref{tab:visors_act}. This sequence leads the eccentricity vector within the deadband (close to the center), and $a \delta \lambda$ below the threshold. Figure \ref{ROE_standby} shows how $a \delta \lambda_{\mathrm{thr}}$ violation drives station-keeping due to the dominating effect of differential drag which causes a decay in relative semi-major axis which couples with Kepler causing a parabolic trajectory of $a \delta \lambda$ \cite{koenig_2017_new, koenig_2018_robust}. This results in a limit cicle in the $a \delta a$–$a \delta \lambda$ plane around the threshold value. Finally, note that the quantization of maneuvers causes a more fragmented behavior in the ROE trajectories.

\subsubsection{Science campaign (transfer to science, science, transfer to standby)}
\begin{figure}[ht!]
    \centering
    {\includegraphics[width=0.7\linewidth]{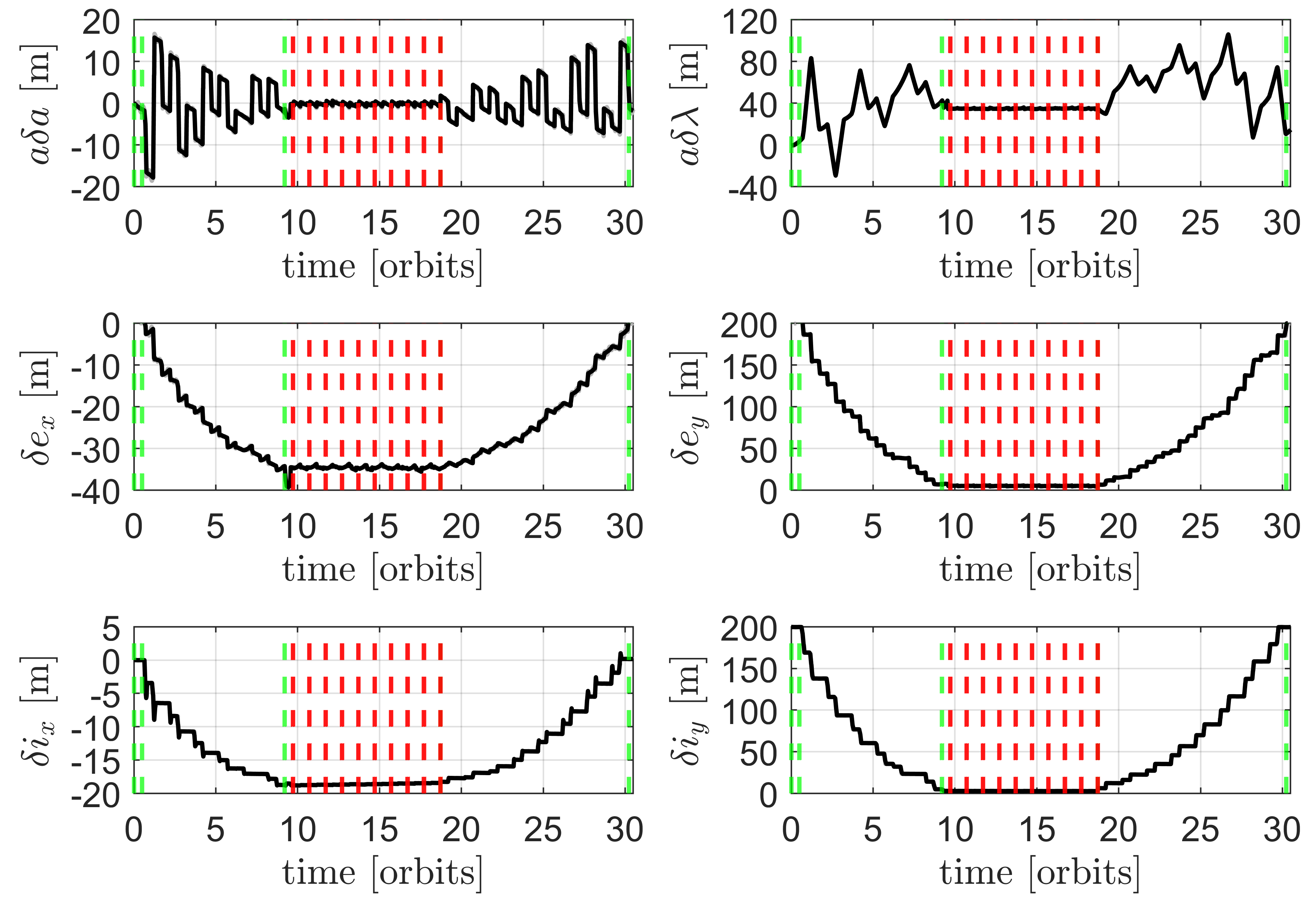}} \\ 
    \vspace{0.3cm}
   {\includegraphics[width=0.36\linewidth]{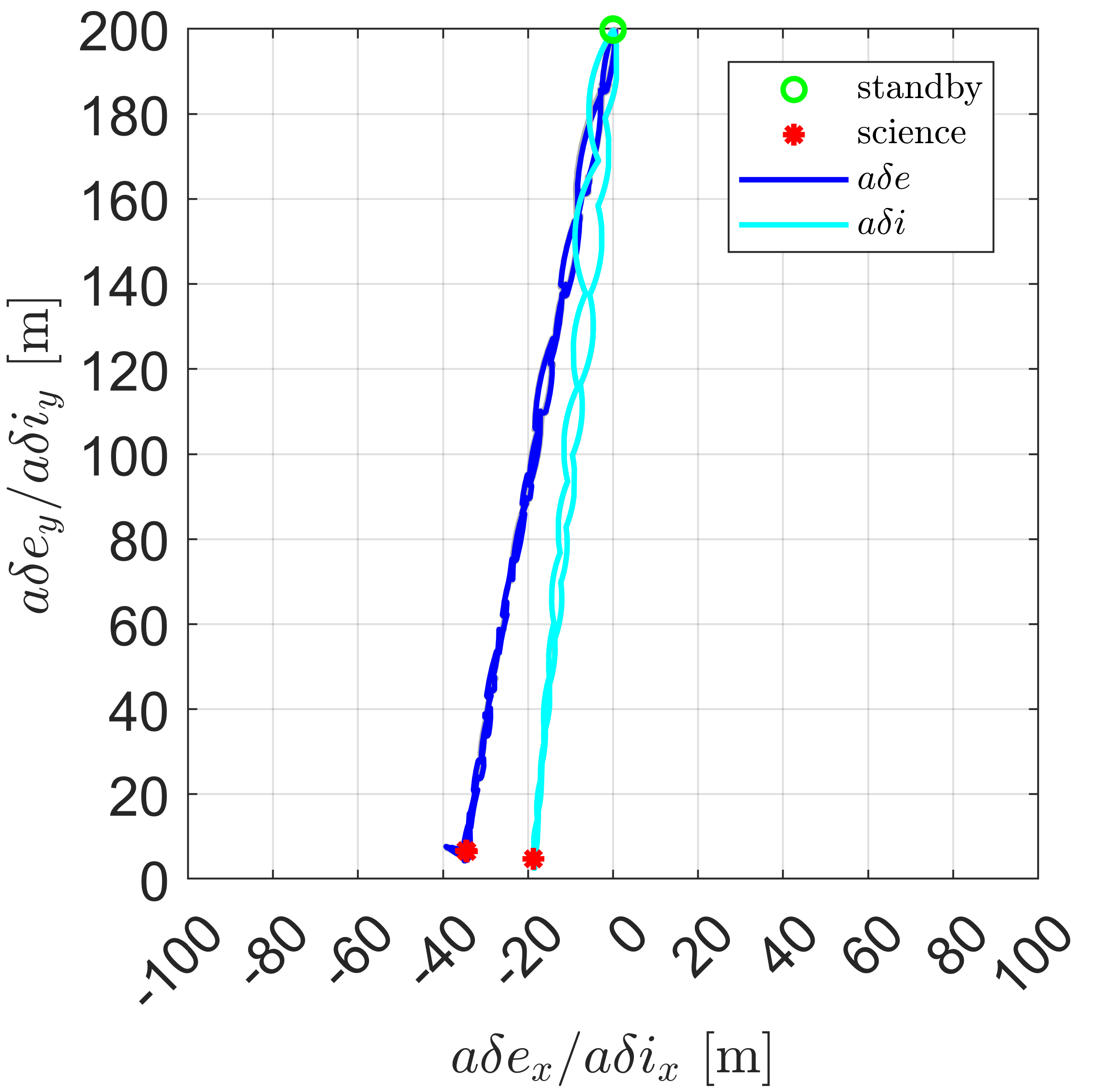}} \hspace{0.5cm}{\includegraphics[width=0.45\linewidth]{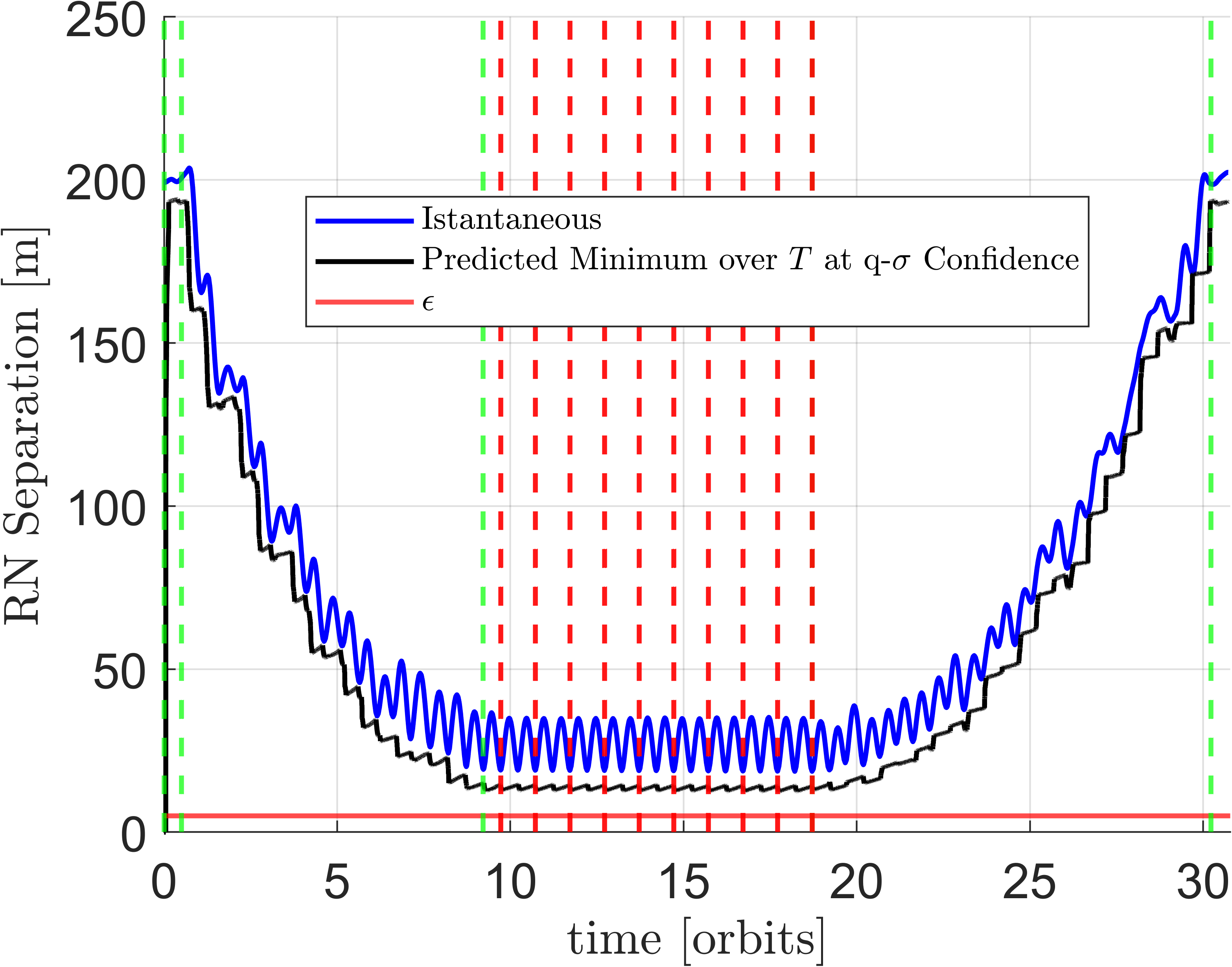}}\caption{Relative orbit elements (ROE) during a science campaign (top six subplots), relative eccentricity and inclination vectors (bottom left), and associated instantaneous and minimum RN separation (bottom right).}\label{ROE_campaign}
\end{figure}

Figure \ref{ROE_campaign} presents the evolution of the ROE during a full science campaign, including transfer to science, science mode (with 10 observation attempts), and transfer to standby modes. During transfer modes, the passively safe guidance profile presented in Section 4 is implemented. The relative eccentricity and inclination vectors move almost in parallel (Figure \ref{ROE_campaign} bottom-left) allowing for passively safe RN separation greater than $\epsilon$ over a horizon of 1.5 h given 3-$\sigma$ uncertainty as verified in real-time by the safety module (Figure \ref{ROE_campaign} bottom-right, where the red vertical dashed lines represent the 10 science observations, and the green vertical dashed lines represent the mode switches). The relative semi-major axis variations scale proportionally to the spacecrat RN separation to avoid compromising the passive safety margin by translating the RN ellipse in radial direction too much (Figure \ref{ROE_campaign} top-left).

\subsubsection{Safety}
\begin{figure}[ht!]
    \centering
    \includegraphics[width=0.45\linewidth]{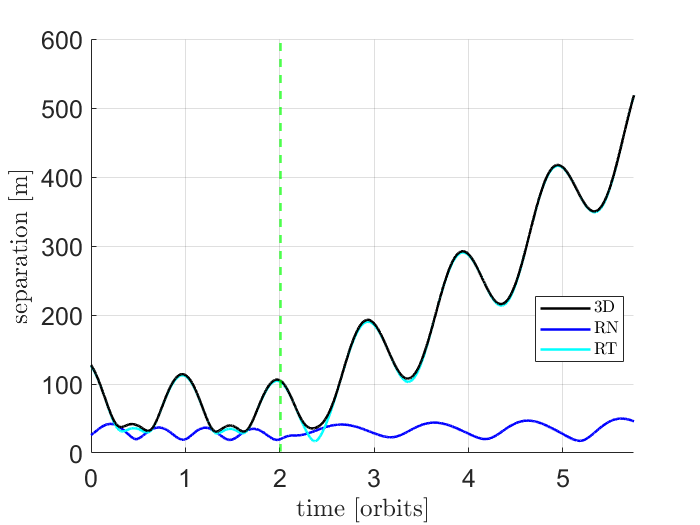}
    \hspace{0.3cm}
    \includegraphics[width=0.45\linewidth]{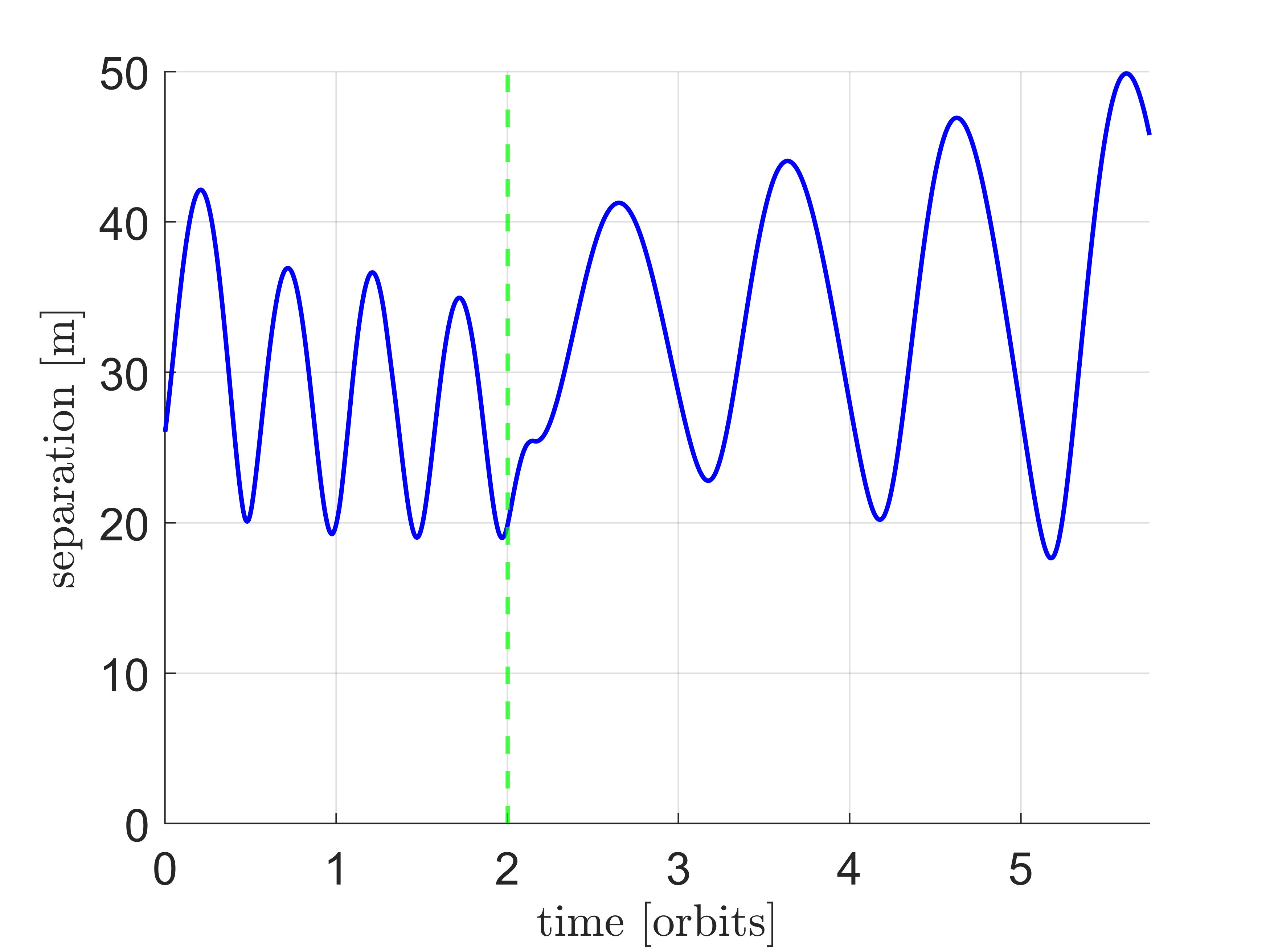}
    \caption{Spacecraft separation in RN, RT, and 3D (left) and zoomed in RN-plane separation (right) before and after an escape maneuver is executed from science campaign}
    \label{fig:escape_sep}
\end{figure}

Figure \ref{fig:escape_sep} shows the evolution of the spacecraft separation in 3D and in the RN and RT planes after executing an escape maneuver computed with alg. \#4 during a science campaign. The relative semi-major axis magnitude is tuned to 5 m, accounting for $3$-$\sigma$ uncertainty of $.036$ m, causing a drift rate of 47 m$\cdot$orbit\textsuperscript{-1} in the along-track direction. Separation above 15 m in the RN plane is maintained for several orbits as the spacecraft drift apart in the along track direction but eventually deteriorates due to drag effects after separation in the along track is established.

\subsubsection{Monte Carlo alignment performances}

\begin{figure}[ht!]
    \centering
    \includegraphics[width=0.3\linewidth]{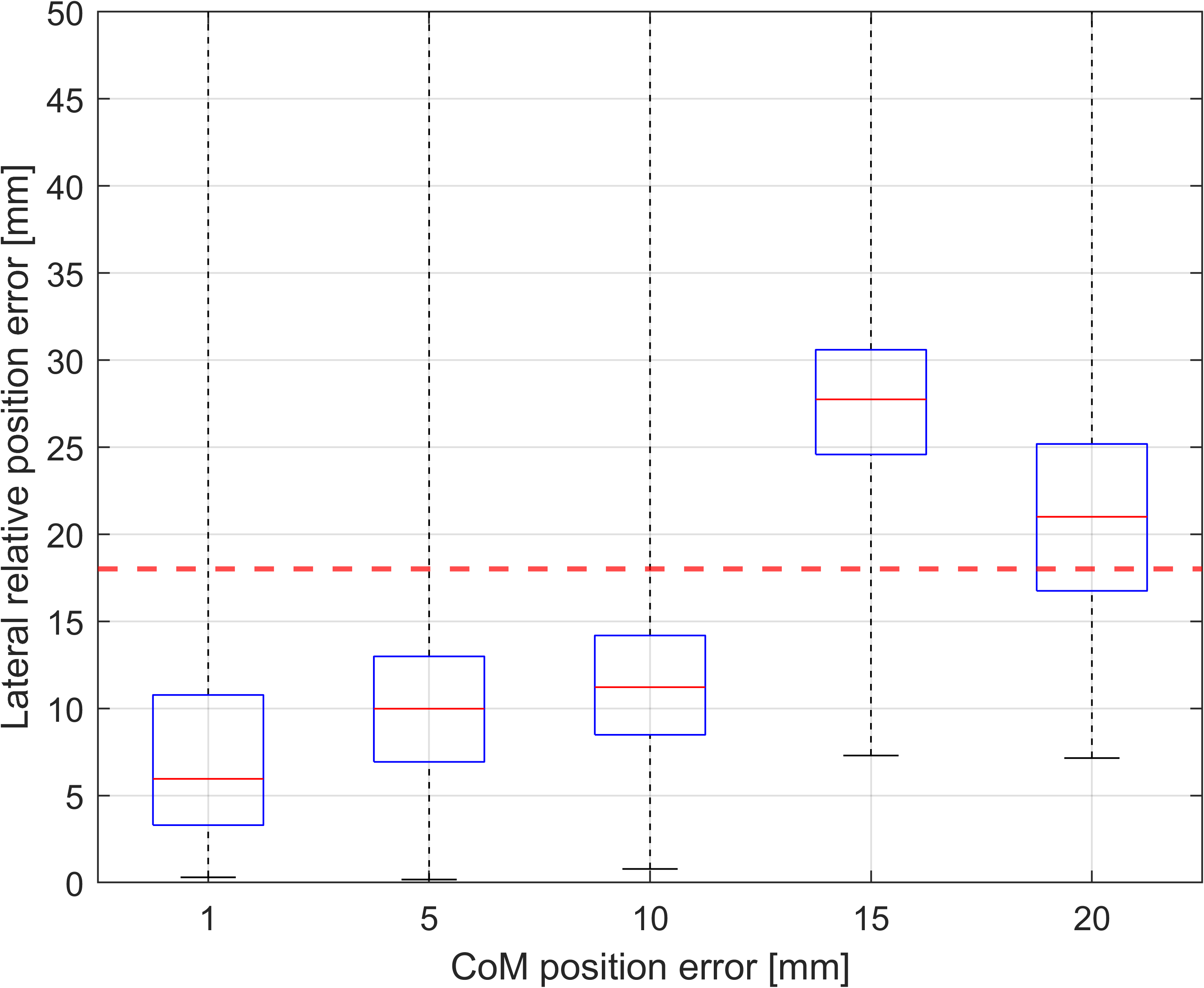} \hspace{0.2cm} \includegraphics[width=0.3\linewidth]{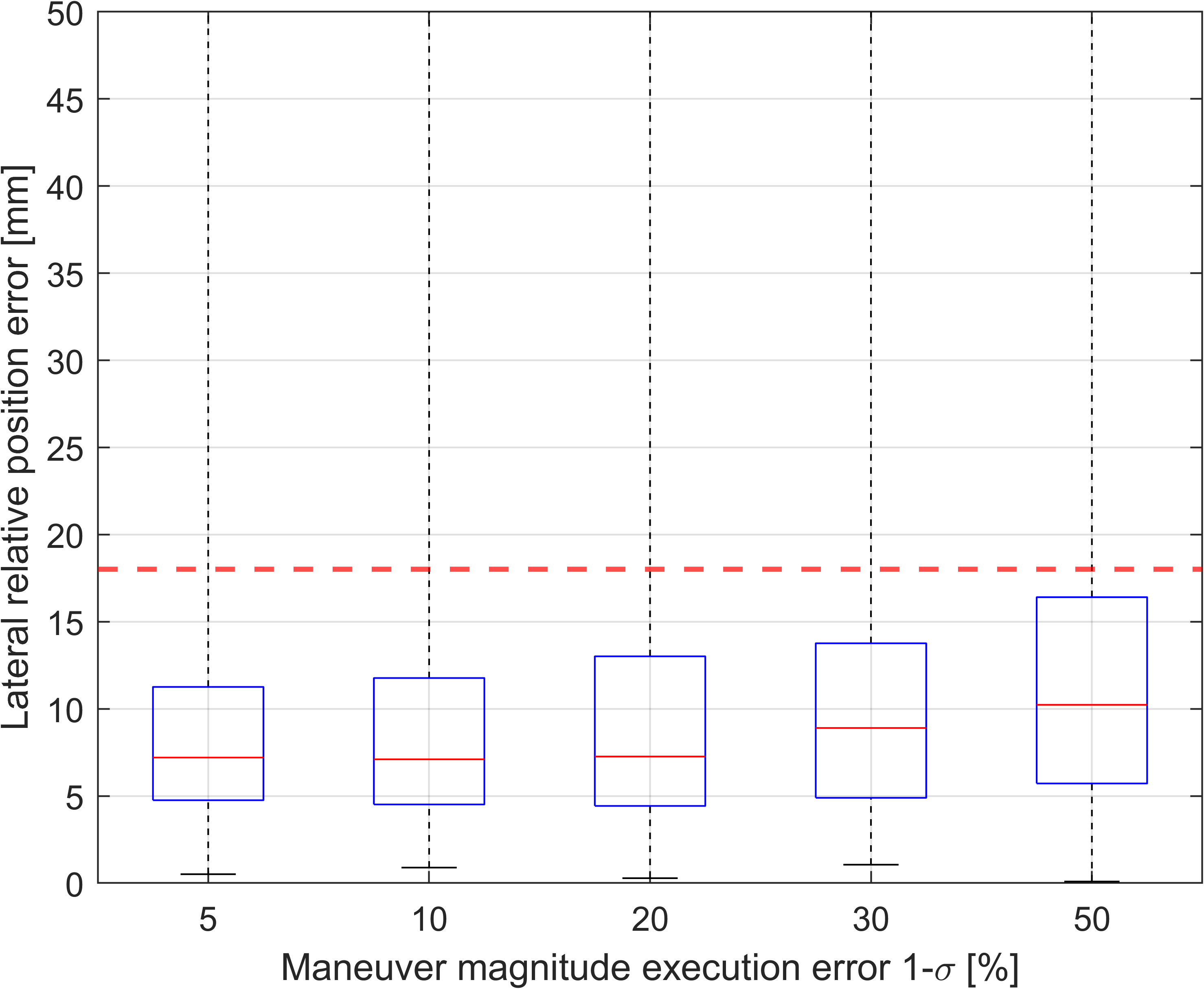}  \hspace{0.2cm} \includegraphics[width=0.3\linewidth]{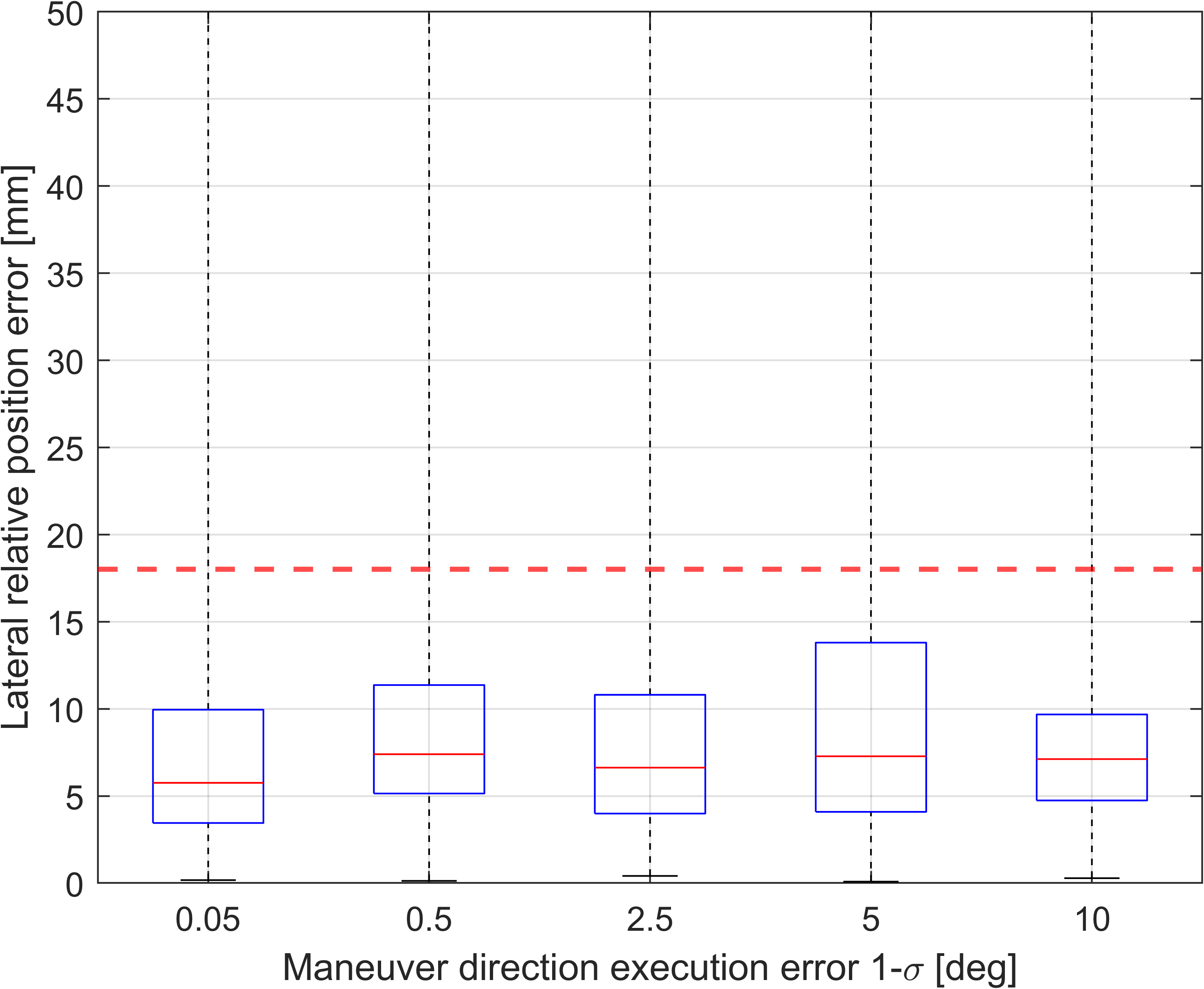} \\
    \vspace{0.3cm}
    \includegraphics[width=0.3\linewidth]{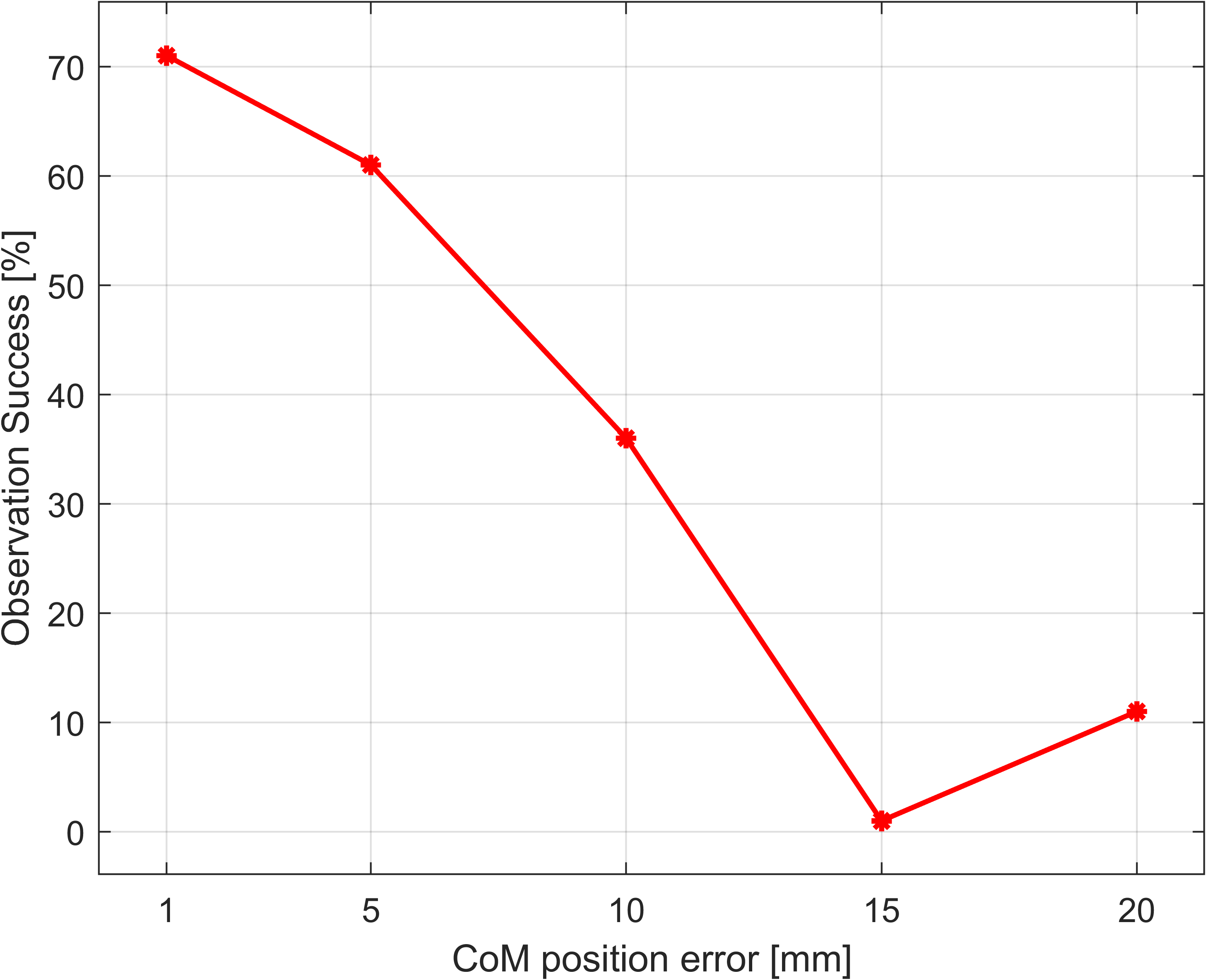}  \hspace{0.2cm}\includegraphics[width=0.3\linewidth]{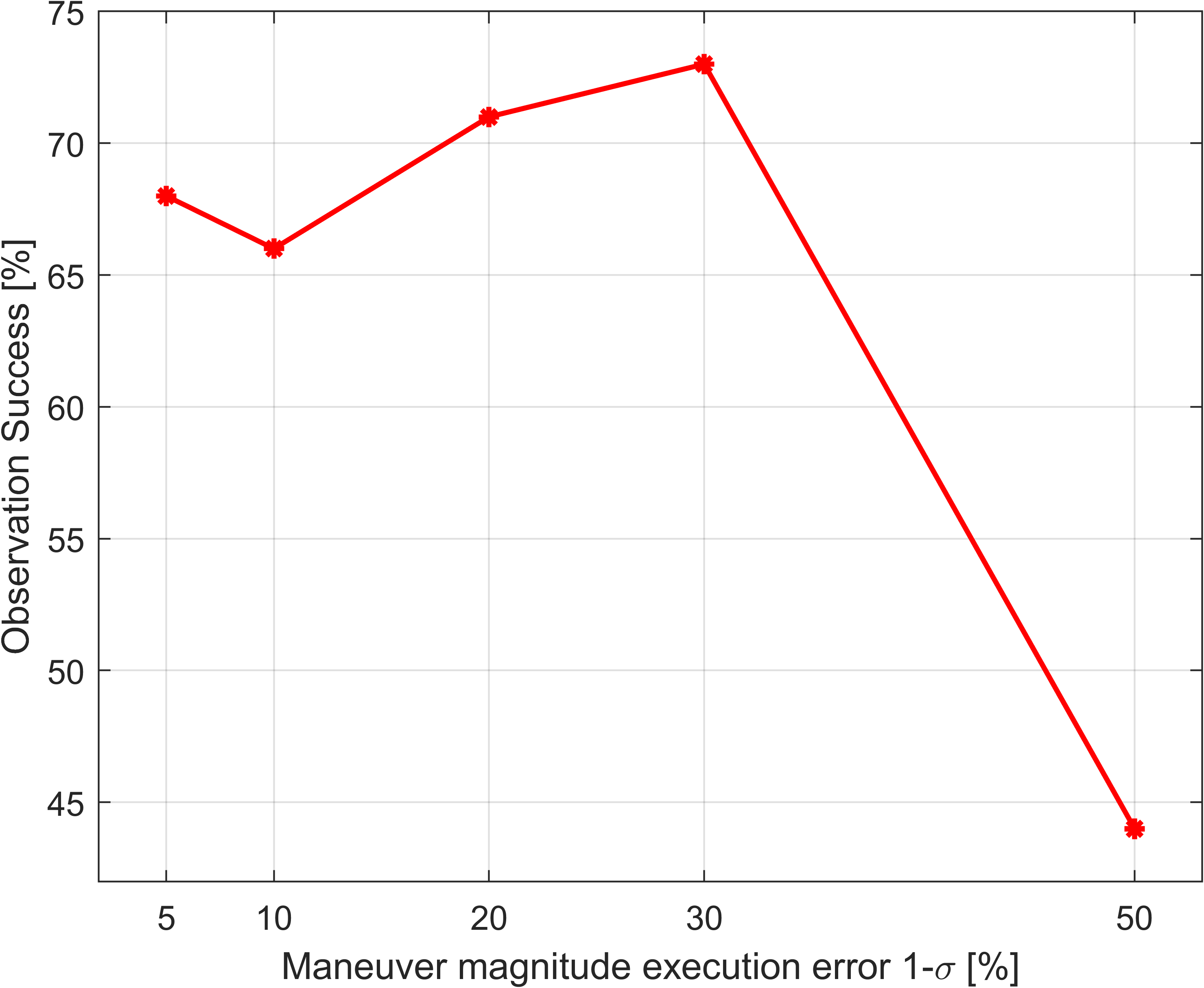}  \hspace{0.2cm}\includegraphics[width=0.3\linewidth]{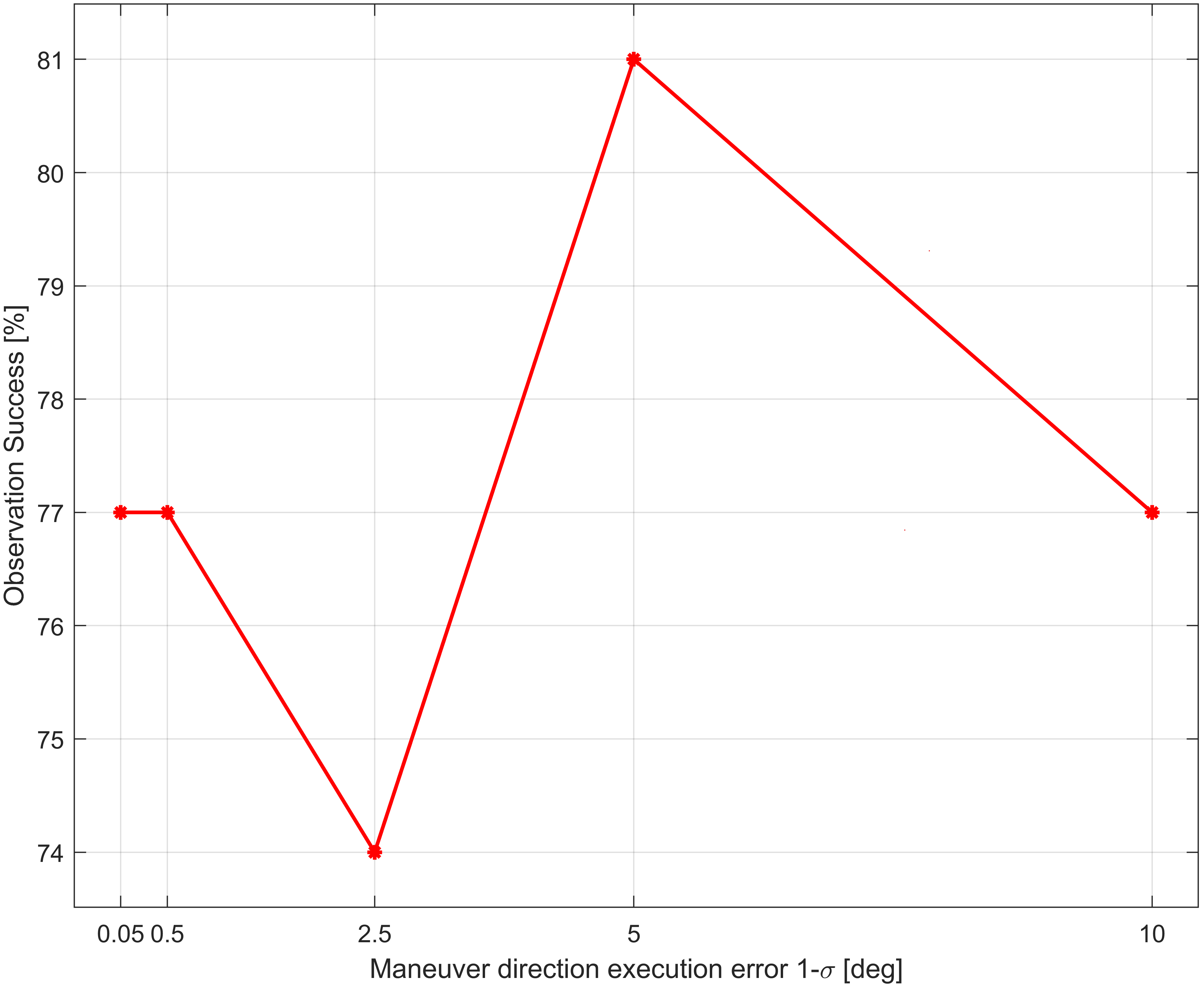} \caption{Alignment metrics (top) and success percentage (bottom) as a function of uncertain parameters.}\label{MC_align_metr}
\end{figure}

Figure \ref{MC_align_metr} shows Monte Carlo results of alignment performance as a function of key uncertain parameters: error in the knowledge of the satellites' center of mass positions (left), and maneuver actuation errors in magnitude (center) and direction (right). For each parameter value ($x$-axis), we present whisker plot statistics of the lateral position alignment error (which tends to be the most challenging requirement to achieve) over 100 observation attempts. Overall, the GNC system achieves the target alignment performances over a broad spectrum of system uncertainties. As on the bottom left, the alignment success percentage is greater than the required 20\% when the center of mass position is known with an accuracy of better than 1 cm. On-ground measurements and on-board calibration should be capable of providing center of mass knowledge of 5 mm, satisfying this requirement. On the bottom center and right, it is shown that the sensitivity of alignment performance to maneuver execution errors is low. The reason is possibly three-fold: (i) in science mode the control loop is closed frequently, from every 5 min (alg. \#2) to every 45 s (alg. \#3), allowing corrections to keep the tracking error small over time; (ii) the last few (1-3) maneuvers have a high impact on the alignment and are generally small for small terminal tracking error before observation, therefore their magnitude error (proportional to the maneuver size, Table \ref{tab:visors_act}) is small and has low impact on alignment performance; (iii) since the execution error is modeled as zero-mean Gaussian, the probability of being close to zero on these few terminal maneuvers is higher, stressing the importance of estimating any non-random biases present in the propulsion system through on-ground and/or on-board calibration.

\section{6. Conclusion} \label{section6}

The Virtual Super-resolution Optics with Reconfigurable Swarms (VISORS) dual CubeSat distributed telescope requires unique performances in terms of high-precision close-proximity formation alignment and fault-tolerant safety for a miniaturized mission of this class, employing low-size-weight-and-power commercial-off-the-shelf technology. This paper shows how the required cm-level accurate alignment at 40 meters separation in low earth orbit can be met safely through a novel GNC architecture composed of: (i) sub-centimeter-level accurate carrier phase differential GPS-based relative navigation with integer ambiguity resolution, (ii) a suite of closed-form and optimization-based impulsive control algorithms providing a broad range of closed-loop control performances, with accuracies from meter-level down to cm-level, over horizons from orbits down to a fraction of an orbit, and (iii) a safety engine integrating passive and reactive safety approaches. Combined navigation and control software-in-the-loop testing reveal the expected performance, surpassing the relative navigation and control requirements for mission success. As a result, alignment guarantees can be probabilistically achieved (\textgreater 20\% observation success rate) by the VISORS GNC architecture in closed loop. In addition, fault-tolerant safety is guaranteed during all mission modes by the proposed architecture. Beyond meeting the VISORS science objectives, these performance results demonstrate the broader impact of this novel GNC architecture, which is itself generalizable to other distributed space missions where high accuracy and fault-tolerant safety are key requirements, such as rendezvous, proximity operations, and swarms.

\section{Appendix}

\subsection{Delta-v budget}\label{app:delta_v_budget}

The $\Delta \boldsymbol{v}$ budget is shown in Table \ref{Tab_dv_budget_VISORS} for all nominal mission modes. This mission plan includes 100 observation attempts divided into ten sets of 10 observations. In between each science campaign the formation spends 2 weeks in standby mode for downlinking. To present a direct metric of fuel consumption the $L_1$ norm of the delta-v expressed in the propulsion thrusters frame is considered (note that the minimization of the $L_2$ norm is explicitly achieved by the optimal control algorithms as in Eq. \ref{OCP_IC}). The mission is fuel positive with a margin that allows extra escapes and re-acquisitions.

\begin{table}[ht!]
\centering
\caption{Delta-v budget}
\begin{tabular}{l l r r}
\hline
\hline
\multicolumn{1}{l}{} & \multicolumn{1}{c}{$|| \Delta \mathbf{v}_{prop} ||_{1}$ (m$\cdot$s\textsuperscript{-1}) } & \multicolumn{1}{l}{\# events} & Total (m$\cdot$s\textsuperscript{-1}) \\
  \cline{1-4}
  Standby tracking (ctrl. alg. \#1) &  0.3488/week & 20 & 6.976\\
  Transfer to science (ctrl. alg. \#1) & 0.4934 & 10 & 4.934 \\
  Transfer to standby (ctrl. alg. \#1) & 0.4844 & 10 & 4.844  \\
  Science tracking (ctrl. alg. \#2+3) & 0.0048 /obs. & 100 & 0.480 \\
  Escape (ctrl. alg. \#4) & 0.0231 & 1 & 0.023 \\
  Acquisition (estimated) & 1.5 & 1 & 1.5 \\
  Re-acquisition (estimated) & 0.5 & 1 & 0.5 \\
  \cline{1-4}
  Total &&& 19.257\\
  Capacity (DSC + OSC) &&& 23.000  \\
  Margin &&& 16.3 \%\\
  \hline
  \hline
\end{tabular}
\label{Tab_dv_budget_VISORS}
\end{table}

\section*{Acknowledgments}
This work is supported by the VISORS mission NSF Award \#1936663. The authors sincerely appreciate the collaborative input from colleagues at the Stanford Space Rendezvous Lab and the Georgia Institute of Technology. The authors also thank NASA Graduate Research Opportunities, and DSO National Laboratories, Singapore, for fellowship support.

\bibliographystyle{AAS_publication}   
\bibliography{main}   

\end{document}